\documentclass[lettersize,journal]{IEEEtran}
\usepackage{amsmath,amsfonts}
\usepackage{algorithmic}
\usepackage{array}
\usepackage[caption=false,font=normalsize,labelfont=sf,textfont=sf]{subfig}
\usepackage{textcomp}
\usepackage{stfloats}
\usepackage{url}
\usepackage{verbatim}
\usepackage{graphicx}
\usepackage[pagebackref=false,breaklinks,colorlinks,bookmarks=false]{hyperref}
\def\BibTeX{{\rm B\kern-.05em{\sc i\kern-.025em b}\kern-.08em
    T\kern-.1667em\lower.7ex\hbox{E}\kern-.125emX}}
\usepackage{balance}

\usepackage{bm}
\usepackage{xcolor}         
\usepackage{multirow}
\usepackage{arydshln}   

\def\eg{\emph{e.g}.} 
\def\ie{\emph{i.e}.} 
\def\cf{\emph{cf}.} 
\def\etc{\emph{etc}} \def\vs{\emph{vs}.}
\definecolor{lightblue}{rgb}{0, 0.65, 0.89}

\newcommand{\revised}[1]{{\color{black}{#1}}}

\newcommand{\graytxt}[1]{{\color{gray}{#1}}}

\newcommand{\lightbluetxt}[1]{{\color{lightblue}{#1}}}

\begin{document}
\title{Generalized Visual Relation Detection with Diffusion Models}
\author{Kaifeng Gao, Siqi Chen, Hanwang Zhang, Jun Xiao$^*$, Yueting Zhuang, and Qianru Sun
\thanks{$^*$Jun Xiao is the corresponding author.}
\thanks{
Kaifeng Gao, Siqi Chen, Jun Xiao, and Yueting Zhuang are with Zhejiang University, Hangzhou, 310027, China. E-mail: kite\_phone@zju.edu.cn, siqic@zju.edu.cn, junx@cs.zju.edu.cn, yzhuang@zju.edu.cn.}
\thanks{Hanwang Zhang is with the School of Computer Science and Engineering, Nanyang Technological University, 639798, Singapore. E-mail: hanwangzhang@ntu.edu.sg.}
\thanks{Qianru Sun is with the School of Information Systems, Singapore Management University, 178902, Singapore. E-mail: qianrusun@smu.edu.sg.}
\thanks{This work was supported by the National Natural Science Foundation of China (62337001) and the Fundamental Research Funds for the Central Universities(226-2024-00058).}
}

\markboth{IEEE TRANSACTIONS ON CIRCUITS AND SYSTEMS FOR VIDEO TECHNOLOGY,~Vol.~*, No.~*, October~2024}%
{How to Use the IEEEtran \LaTeX \ Templates}

\maketitle

\begin{abstract}
    Visual relation detection (VRD) aims to identify relationships (or interactions) between object pairs in an image. Although recent VRD models have achieved impressive performance, they are all restricted to pre-defined relation categories, while failing to consider the \emph{semantic ambiguity} characteristic of visual relations. Unlike objects, the appearance of visual relations is always subtle and can be described by multiple predicate words from different perspectives, e.g., ``ride'' can be depicted as ``race'' and ``sit on'', from the sports and spatial position views, respectively. To this end, we propose to model visual relations as continuous embeddings, and design diffusion models to achieve generalized VRD in a conditional generative manner, termed Diff-VRD. We model the diffusion process in a latent space and generate all possible relations in the image as an embedding sequence. During the generation, the visual and text embeddings of subject-object pairs serve as conditional signals and are injected via cross-attention. After the generation, we design a subsequent matching stage to assign the relation words to subject-object pairs by considering their semantic similarities. 
    Benefiting from the diffusion-based generative process, our Diff-VRD is able to generate visual relations beyond the pre-defined category labels of datasets. 
    To properly evaluate this generalized VRD task, we introduce two evaluation metrics, \ie, text-to-image retrieval and SPICE PR Curve inspired by image captioning. Extensive experiments in both human-object interaction (HOI) detection and scene graph generation (SGG) benchmarks attest to the superiority and effectiveness of Diff-VRD.
\end{abstract}


\section{Introduction}

\begin{figure}
  \centering
  \includegraphics[width=\linewidth]{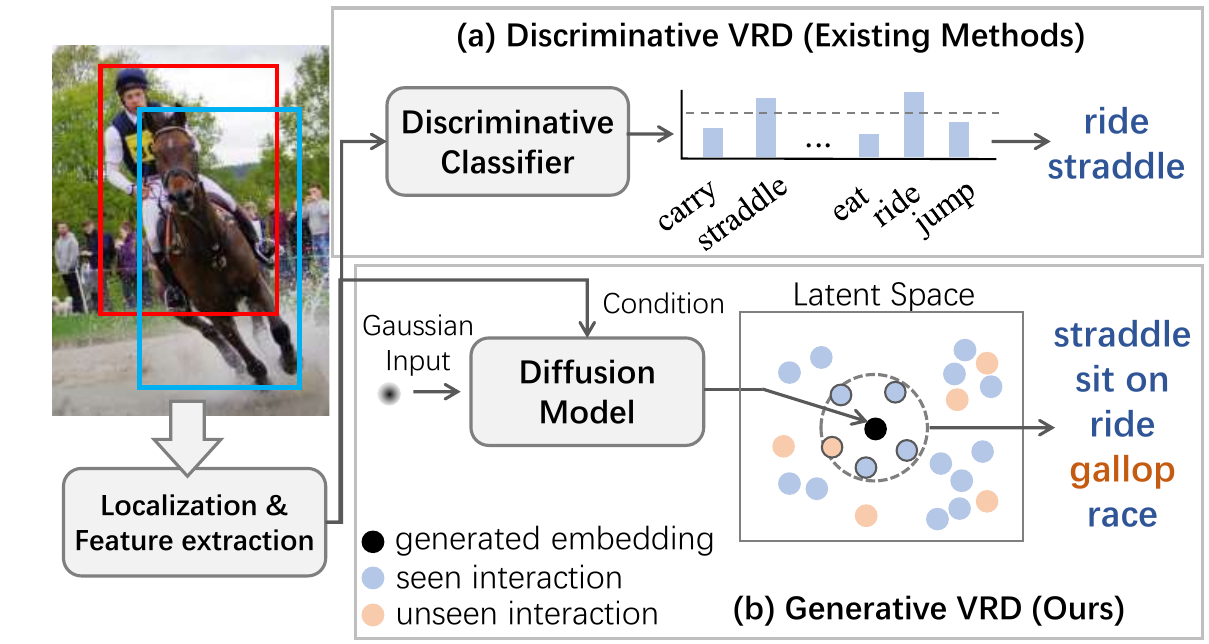}
  \caption{An example of visual relation detection (VRD). (a) Existing methods achieve VRD through discriminative classification. (b) We propose to generate visual relations via a diffusion model, conditioned on visual features of subject-object pairs. We model all visual relations collectively as one sequence but display only the relations of one subject-object pair for clarity.%
  }\label{fig_fig1}
\end{figure}

Visual Relation Detection (VRD) aims to localize subject-object pairs and recognize pairwise relations (or interactions), such as the ``person-ride-horse'' example illustrated in Figure~\ref{fig_fig1}. It serves as a fundamental component for AI agents to comprehend scenes and perceive their surroundings. Recent VRD models have demonstrated impressive performance by leveraging state-of-the-art model designs: either in an end-to-end manner~\cite{cong2023reltr,li2022sgtr,wang2024ted,zhang2021mining,zhong2022towards,cheng2023multi} based on DETR (DEtection TRansformer) frameworks \cite{carion2020end}, or in a two-stage fashion~\cite{zhang2022efficient,ren2024learning,li2023label,fu2023drake,wang2024novel,tian2024gaussian}, where subject-object pairs are initially localized, followed by relation recognition.

However, a significant limitation of existing approaches lies in their constrained definition of relations within predefined category labels. This limitation poses challenges when attempting to extend the vocabulary of relations to include new categories. Such extension typically involves labor-intensive data annotation for new categories and computationally expensive retraining or finetuning of models. Consequently, an ideal solution would be training VRD models capable of generalized visual relation detection. Such a model should be trained on all available relation labels and be able to conjecture potential new relation categories at test time.

Similar attempts have been widely explored in the field of object detection. Existing works enable detectors to be trained on limited annotations while being able to recognize new object categories at test time (\eg, open-vocabulary object detection~\cite{zareian2021open,gu2022open,wu2023cora}). They classify unseen objects into specific categories by leveraging external knowledge transferred from pre-trained visual-language models (VLMs) such as CLIP \cite{radford2021learning}. However, addressing visual relations presents a more significant challenge due to their inherent \textbf{semantic ambiguity}. Unlike objects, the appearance of relations is often subtle and can be described by multiple predicate words from different perspectives, \eg, the relation of person and horse in Figure~\ref{fig_fig1} can be described as ``ride'', ``straddle'', and so on. Therefore, recognizing visual relations by \emph{discriminative} classification might be sub-optimal. Although zero-shot~\cite{wu2023end} or open-vocabulary~\cite{he2022towards,li2024pixels,zhang2023learning} settings are also explored in existing VRD methods, they still resort to \emph{discriminative} multi-label classification, as shown in Figure~\ref{fig_fig1}~(a), which treats each interaction discretely and ignores the semantic ambiguity.

To this end, we propose to model visual relations using continuous embeddings, as depicted in Figure~\ref{fig_fig1}~(b), and perform generalized relation detection in a  \emph{generative} manner. 
Considering the inherent interconnectedness of visual relations (or interactions)\footnote{we use the words ``visual relation'' and ``interaction'' interchangeably depending on the context.}, we treat all interactions in the image holistically to capture overall contexts and develop a diffusion model to ``decode'' these interactions in a generative manner. This diffusion model, dubbed Diff-VRD, is specially tailored for visual relation detection.
It operates by using a sequence of Gaussian noise vectors as input, which are gradually denoised to generate interaction embeddings conditioned on the visual cues of subject-object pairs. The diffusion process is modeled using a transformer decoder, and cross-attention layers are employed to incorporate the conditional signals from visual appearance. The underlying working logic is that the presence of Gaussian noise in the interaction embeddings forces the model to consider semantic ambiguity (during denoising). Meanwhile, as the conditional signals are introduced, the interaction embeddings are learned to be semantically meaningful by aligning with the visual appearance. 

Given the fact that visual relations would be eventually described by discrete words (or phrases), we extend the diffusion-denoising process with an embedding step and a rounding step (Sec.~\ref{sec_vlb}), in which an embedding function is jointly trained with the diffusion model as in ~\cite{li2022diffusion,gong2023diffuseq}. 
After generating the relation sequence, we conduct a matching step to match each relation in the sequence to subject-object pairs according to semantic similarities (Sec.~\ref{sec_grounding}). In addition, we design an auxiliary loss and introduce matching supervision into the diffusion model, which empirically improves the matching quality 
(Sec.~\ref{sec_grounding_superv}). 
Thanks to the generative process in the diffusion model, our Diff-VRD is able to identify visual relations that are not covered by the pre-defined categories in datasets, \ie, achieving generalized VRD by rounding the relation embeddings to discrete phrases according to a large external relation vocabulary. 

To effectively evaluate the generalized VRD, we introduce two proxy tasks and metrics: namely text-to-image (T2I) retrieval and SPICE~\cite{anderson2016spice} precision-recall curve (PR curve) adapted from image captioning. They account for results that go beyond the ground-truth annotations, capturing reasonable but unannotated relation categories.
The experiment results show that our Diff-VRD outperforms conventional state-of-the-art VRD models on both T2I retrieval and SPICE. For example, compared to UPT~\cite{zhang2022efficient}, we achieve an absolute improvement of 11.31\% Recall@10 in terms of T2I retrieval on the HICO-DET~\cite{du2022learning} dataset. 
\revised{We also evaluated the relation diversity using the SPICE targets. The results show that, compared to existing methods—both the closed-set approach UPT~\cite{zhang2022efficient} and the open-set
methods GNE-VLKT~\cite{liao2022gen} and the CLIP~\cite{radford2021learning} baseline, Diff-VRD correctly identifies a larger number of meaningful relations.
}
In addition, Diff-VRD can also be applied on top of existing VRD models as an enhancement module (\ie, post-processing) to refine the relation predictions. 
When applied on IEtrans~\cite{zhang2022fine}, it brings an absolute improvement of 5.60\% recall of the T2I retrieval task on the Visual Genome~\cite{krishna2017visual} dataset.
%


Our contributions are thus three-fold. 1) We achieve generalized visual relation detection in a generative paradigm for the first time,
which considers the inherent semantic ambiguity of visual relations. 2) We 
introduce two new metrics (\ie,  text-to-image retrieval and SPICE PR curve) to properly evaluate reasonable visual relations that go beyond the ground-truth annotations. 3) We propose a diffusion-based VRD model, \ie, Diff-VRD. It achieves comparable performance with SOTA VRD models while excelling in capturing more subtle and diverse visual relationships.


\section{Related Works}

\textbf{Visual Relation Detection (VRD)} has been widely studied in tasks such as scene graph generation (SGG)~\cite{xu2021scene,li2022sgtr,cong2023reltr,fu2023drake} and human-object detection (HOI)~\cite{yang2021rr,wang2024ted,zhang2022efficient,cheng2023multi,park2023viplo}. Existing works on VRD can be grouped into two-stage methods and one-stage (end-to-end) methods. Two-stage methods~\cite{zhao2022semantically,li2023label,ren2024learning,zhang2022efficient,park2023viplo} first apply an off-the-shelf object detector to detect object bounding boxes, and then exhaustively combine them as subject-object pairs which are fed to a subsequent module for interaction classification. One-stage methods~\cite{cong2023reltr,li2022sgtr,zhang2021mining,cheng2023multi,zhong2022towards} detect subject-object bounding boxes and their visual relations simultaneously, typically in a DETR~\cite{carion2020end}-based paradigm. They adopt a set of learnable queries as the input of the Transformer decoder, aggregate the image-wide context information based on the cross-attention layers, and finally predict the bounding boxes simultaneously with their pair-wise relations. Unlike existing works that classify visual relations within pre-defined categories, we recognize visual relations in a generative paradigm by diffusion models, covering relation categories that go beyond the ground-truth annotations.

\textbf{Zero-Shot and Open-vocabulary VRD} aim to recognize new relation categories which are unseen during training. A straightforward and feasible way to achieve Zero-Shot VRD is aligning visual representations to pre-trained word embeddings (\eg, GloVe~\cite{pennington2014glove} or ELMo~\cite{peters2018deep}), and classifying the visual relations according to the similarities between them~\cite{wang2018zero,liu2020consnet,bansal2020detecting}. Recently, pretrained visual-language models (VLMs) have demonstrated their extraordinary zero-shot ability (\eg, CLIP~\cite{radford2021learning}, X-VLM~\cite{zeng2022multi}) by learning a joint visual-language space through contrastive pretraining~\cite{chuang2020debiased}. Consequently, many works adopted such VLMs and transferred the knowledge of the learned joint space to their VRD models~\cite{wu2023end,he2022towards,li2024pixels,zhang2023learning}. With the encyclopedic knowledge from VLMs, they can achieve \emph{open-vocabulary} VRD when an open-world relation vocabulary is provided at deployment. However, they (both zero-shot and open-vocabulary VRD) still recognize visual relations by \emph{discriminative} classification, and evaluate the results on seen/unseen (also referred to as base/novel) relation categories split within the annotations of datasets. In contrast, Diff-VRD is the first one to recognize visual relations in a \emph{generative} manner, and introduce new proxy tasks to properly evaluate reasonable but unannotated relation categories~\footnote{Our model can generate relations not only go beyond the annotations of the train set but also the test set. Hence our setting is more general than the base/novel category splits of the existing open-vocabulary setting.}.

\textbf{Diffusion Models} are generative models that aim to model a target distribution $\bm{x}_0 \sim q(\bm{x})$ by learning a denoising process $p_{\theta}(\bm{x}_{t-1} | \bm{x}_t)$ with varying noise levels. Conditional generation can be achieved by introducing classifier-free guidance~\cite{ho2021classifier} to the denoising process as $p_{\theta}(\bm{x}_{t-1} | \bm{x}_t, \bm{y})$, where the conditional signal $\bm{y}$ can be injected via cross-attention~\cite{rombach2022high}, token-wise concatenation~\cite{gong2023diffuseq}, adaptive layer normalization~\cite{lu2024vdt}, \etc. 
Diffusion models have demonstrated remarkable success in vision~\cite{ho2020denoising,rombach2022high}, audio~\cite{kong2020diffwave,kim2022guided}, and language~\cite{li2022diffusion,gong2023diffuseq} domains. Our work for the first time introduces diffusion models to visual relation detection, and achieves generalized VRD beyond the datasets' annotations.


\revised{
\textbf{Diffusion-based Discrete Text Generation}. Recent works have explored the application of diffusion models for discrete text generation in vision-language tasks, \eg, image and video captioning.
%
Bit Diffusion~\cite{chen2023analog} encodes the text tokens into binary bits and then represents them as real numbers (analog bits), allowing the model to follow a continuous diffusion process. Following this spirit, SCD-Net~\cite{luo2023semantic} concatenates semantic features to the analog bits and encodes them via self-attention. MM-Diff-Net~\cite{kainulainen2024diffusion} explores various modality fusion techniques to incorporate multi-modal semantic features.
In addition, some works~\cite{zhu2022exploring,tian2024image} use discrete masks instead of continuous noise to model the diffusion process and develop image captioning models.
We follow the continuous diffusion process. But unlike Bit Diffusion~\cite{chen2023analog}, we introduce the embedding-rounding module. It projects the discrete relation words to continuous embeddings and initializes them with CLIP text features, which is inherently more sematic meaningful than directly converting text tokens into binary bits. 
}

\section{Method}

\begin{figure*}
  \centering
  \includegraphics[width=\linewidth]{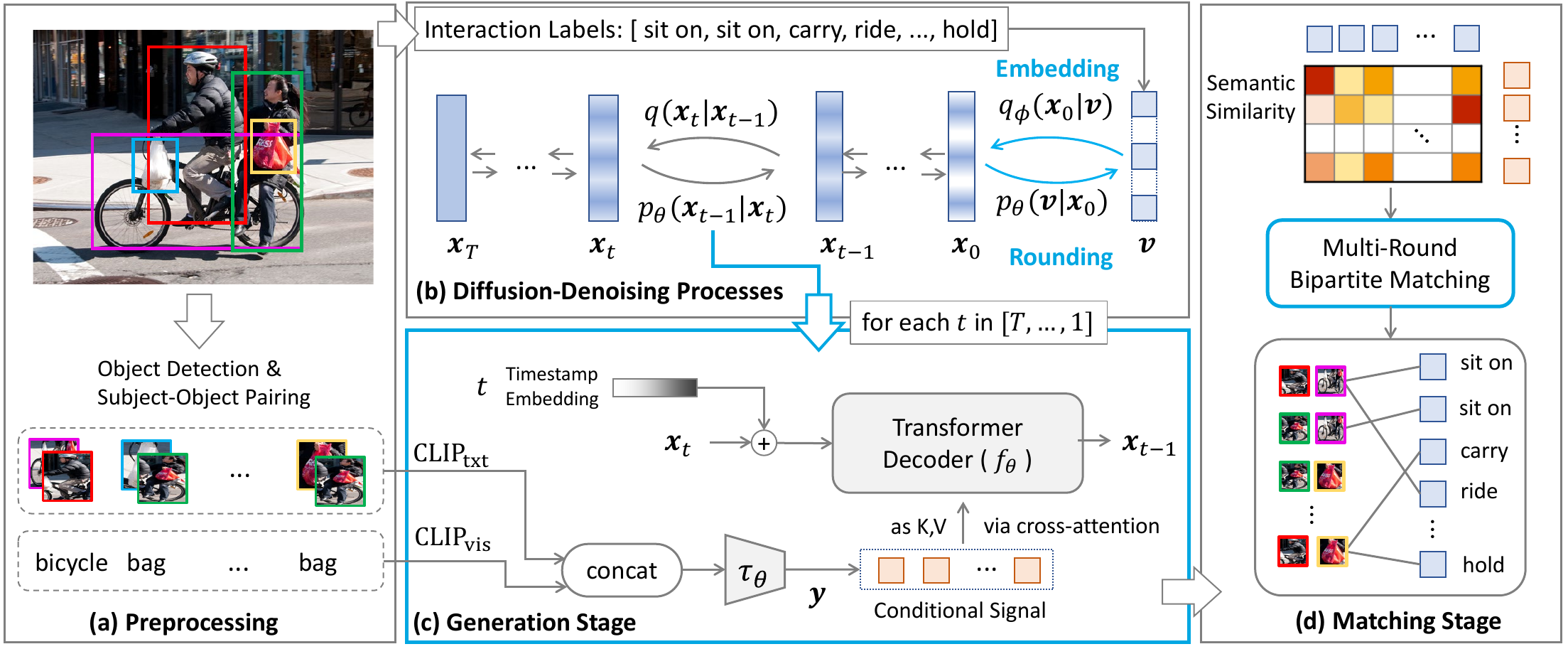}
  \caption{\revised{Overview of Diff-VRD. (a): Given an image, we first detect all the objects with their categories and enumerate the subject-object pairs. (b): We extend the diffusion-denoising process with an embedding step $q_\phi(\bm{x}_0 | \bm{v})$ and a rounding step $p_\theta(\bm{v} | \bm{x}_0)$. We introduce a tailored training objective for this extended diffusion-denoising process, as described in Sec.~\ref{sec_vlb}. (c): The parameter $\theta$ in the denoising process is implemented by a Transformer decoder $f_\theta$. The conditional signal includes CLIP encoded visual and text embeddings of subject-object pairs, which are concatenated as $\bm{y}$, projected by $\tau_\theta$, and injected via cross-attention layers (\cf~Eq.~(\ref{eq:crossattn})). (d): After generating the relation sequence $\bm{v}$, we assign each $v_i$ to a subject-object pair by using multi-round bipartite matching based on their semantic similarities. Our contributions are highlighted in \lightbluetxt{light blue}.}
}\label{fig_pipeline}
    \vspace{-2ex}
\end{figure*}
We first briefly introduce the preliminaries of diffusion models and the problem setup of generalized VRD (Sec.~\ref{sec_Preliminaries}). Then, we introduce the proposed Diff-VRD, as illustrated in Figure~\ref{fig_pipeline}. Diff-VRD first generates a relation sequence conditioned on subject-object features via diffusion models (Sec.~\ref{sec_vlb}), and then matches visual relations to certain subject-object pairs based on their similarities (Sec.~\ref{sec_grounding}). Finally, we introduce auxiliary matching supervision into Diff-VRD (Sec.~\ref{sec_grounding_superv}) to improve the matching quality.


\subsection{Preliminaries and Problem Setup}\label{sec_Preliminaries}
\textbf{Preliminaries}. Diffusion models~\cite{ho2020denoising,song2021denoising} 
are designed to model a target distribution $\bm{x}_0 \sim q(\bm{x}) $ by learning a denoising (reverse) process with arbitrary noise levels. To do this, a Markov diffusion (forward) process is defined to gradually corrupt $\bm{x}_0$ into a standard Gaussian noise. 
\revised{Specifically, each forward step samples $\bm{x}_t$ as 
\begin{equation}\label{eq:q_sample}
    \bm{x}_t \sim q(\bm{x}_t | \bm{x}_{t-1}) = \mathcal{N}(\bm{x}_t;\sqrt{1-\beta_t}\bm{x}_{t-1},\beta_t\bm{I}),
\end{equation}
where $t=1,\ldots,T$ and $\beta_t \in (0,1)$ is the variance schedule. 
By applying the reparameterization trick~\cite{ho2020denoising}, $\bm{x}_t$ can also be directly sampled conditioned on $\bm{x}_0$, \ie, 
$\bm{x}_t = \sqrt{\bar{\alpha}_t}\bm{x}_0 + \sqrt{1-\bar{\alpha}_t}\bm{\epsilon}_t$, 
where $\bm{\epsilon}_t \sim \mathcal{N}(\bm{0},\bm{I})$ and $\bar{\alpha}_t = \prod_{i=1}^t(1-\beta_i)$.}
Given the forward process, a diffusion model is then trained to approximate the denoising process, \revised{where each step samples
\begin{equation}
\bm{x}_{t-1} \sim p_\theta(\bm{x}_{t-1} | \bm{x}_t) = \mathcal{N}(\bm{x}_{t-1};\bm{\mu}_\theta(\bm{x}_t,t),\bm{\Sigma}_\theta(\bm{x}_t,t)),    
\end{equation}}
where $\theta$ contains learnable parameters, \eg, characterized by a U-Net~\cite{rombach2022high} or a Transformer~\cite{peebles2023scalable}.

\revised{
Existing diffusion models~\cite{ho2020denoising,nichol2021improved,peebles2023scalable} are trained with the variational lower bound (VLB) of $\bm{x}_0$'s log-likelihood, which can be formulated as 
\begin{align}\label{eq_ori_vlb}
  \mathcal{L}_{\text{VLB}}^\text{ori} (\bm{x}_0) &= 
  \mathop{\mathbb{E}}_{\bm{x}_{1:T} \sim q(\bm{x}_{1:T} | \bm{x}_0)} \left[
    \log \frac{q(\bm{x}_T | \bm{x}_0)}{p(\bm{x}_T)}
    \right. \notag \\
    & \left. + \sum_{t=2}^T \log \frac{q(\bm{x}_{t-1} | \bm{x}_t,\bm{x}_0)}{p_\theta(\bm{x}_{t-1}|\bm{x}_t)}
    - \log p_\theta(\bm{x}_0 | \bm{x}_1)
  \right].
\end{align}
Here the first two terms are the the Kullback-Leibler (KL) divergence of $q$ and $p_\theta$. Since $q$ and $p_\theta$ are both Gaussian, their KL divergence is determined by the mean $\bm{\mu}_\theta$ and covariance $\bm{\Sigma}_\theta$. Following~\cite{ho2020denoising}, a common approach is to define $\bm{\Sigma}_\theta$ as a fixed variance schedule, \ie,  $\bm{\Sigma}_\theta=\sigma_t^2\bm{I}$, where $\sigma_t^2=\frac{1-\bar{\alpha}_{t-1}}{1-\bar{\alpha}_{t}}\beta_t$. Then, 
By re-parameterizing $\bm{\mu}_\theta$ as a noise prediction network, the model can be optimized using a simplified training objective:
\begin{align}\label{eq_ori_simple}
  \mathcal{L}_{\text{simple}}^\text{ori}(\bm{x}_0)=
  \mathop{\mathbb{E}}_{\bm{x}_t,\bm{\epsilon},t}
  \left[ \|\bm{\epsilon}_\theta(\bm{x}_t,t) - \bm{\epsilon}\|_2^2\right],  
\end{align}
where $\bm{x}_t \sim q(\bm{x}_t | \bm{x}_0), \bm{\epsilon} \sim \mathcal{N}(0,1)$, and $t \sim \mathcal{U}\{1,2,...,T\}$. 
%
}

\textbf{Problem Setup}.  
Given an image, the task of VRD aims to detect all the objects and identify their pair-wise visual relations. In our work, we follow the two-stage VRD paradigm. We first apply an off-the-shelf detector to obtain a set of object predictions, where each prediction is characterized by a bounding box and an object category $c_o \in \mathcal{C}_o$. Then, we recognize all the possible visual relations (or interactions) for each subject-object pair $(i,j)$, denoted as $\mathcal{R}_{i,j} =  \{r_1,\ldots,r_k\} \subseteq \mathcal{C}_r$. 
Here $\mathcal{C}_o$ and $\mathcal{C}_r$ are sets of object and relation categories defined by the annotations from VRD datasets. In our setting, we replace $\mathcal{C}_r$ with a much larger vocabulary $\mathcal{V}$ (\ie, $\mathcal{C}_r \subseteq \mathcal{V}$) to achieve generalized VRD~\footnote{In practice, $\mathcal{V}$ contains phrases and prepositions such as ``sit on'', ``in front of''. Refer to Sec.~\ref{sec_implementation} for details.}. Since our main focus is relation detection, we keep the object category set $\mathcal{C}_o$ unchanged for simplicity. 


\subsection{Conditional Relation Generation via Diffusion Models}\label{sec_vlb}

We first extend the vanilla diffusion pipeline to do relation generation, and then introduce the conditional information based on the subject-object pairs, as shown in Figure~\ref{fig_pipeline}.

\textbf{Relation Generation}. Given the fact that relation categories are described by discrete text, we must map the text into a continuous space at the beginning of the forward process, and round the continuous embedding back into text at the end of the reverse process. 
Following \cite{li2022diffusion,gong2023diffuseq}, we design an embedding function $\text{Emb}_\phi$ to map each $v_i \in \mathcal{V}$ to a continuous vector in $\mathbb{R}^d$. Then, we concatenate all the relations in the image as a sequence $\text{Emb}_\phi(\bm{v}) = [\text{Emb}_\phi(v_1),\ldots,\text{Emb}_\phi(v_L)] \in \mathbb{R}^{L\times d}$, where $L$ is the total number of relations in the image ($L$ is set to a fixed value that is large enough during testing).


\revised{
Consequently, we extend the forward process with an extra Markov transition from the discrete relation sequence $\bm{v}$ to $\bm{x}_0$ (as shown in Figure~\ref{fig_pipeline}~(b)), parameterized by $q_\phi(\bm{x}_0 | \bm{v}) = \mathcal{N}(\bm{x}_0;\text{Emb}_\phi(\bm{v}),\sigma_0\bm{I})$.} On the other hand, the reverse process is extended with a trainable rounding step, parameterized by $p_\theta(\bm{v} | \bm{x}_0) = \prod_{i=1}^L p_\theta(v_i | \bm{x}_0^{(i)})$, \revised{where $v_i$ and $\bm{x}_0^{(i)}$ are the $i$-th item in $\bm{v}$ and $\bm{x}_0$, respectively.}

\revised{A closer analysis reveals that the embedding-rounding step functions like an extra variational auto-encoder (VAE)~\cite{kingma2013auto} hierarchy~\footnote{\revised{Diffusion models can be viewed as a special type of hierarchical VAEs that has multiple hierarchies over latent variables~\cite{luo2022understanding,kingma2021variational,sonderby2016ladder,kingma2016improved}}} 
on top of $\bm{x}_0$.}
Thus, the training objective of our model can be derived by incorporating the original VLB in Eq.~(\ref{eq_ori_vlb}) and the following VAE training objective:
\begin{align}\label{eq:vae}
  \min_{\theta,\phi} 
  \mathop{\mathbb{E}}_{q(\bar{\bm{z}} | \bar{\bm{x}} )}\left[
    \log \frac{q_\phi(\bar{\bm{z}}| \bar{\bm{x}})}{p_\theta(\bar{\bm{x}} | \bar{\bm{z}})p(\bar{\bm{z}})}
  \right].
\end{align}
\revised{Note that Eq.~(\ref{eq:vae}) denotes a generic VAE, where we use  $\bar{\bm{x}},\bar{\bm{z}}$ to represent the variables to avoid confusion.}
By comparing the embedding-rounding step w.r.t. VAE, it can be observed that $\bm{x}_0$ serves as the latent variable  $\bar{\bm{z}}$ and $\bm{v}$ serves the variable $\bar{\bm{x}}$. Besides, since $\bm{x}_0$ is further conditioned on $\bm{x}_1$, yielding that $p(\bar{\bm{z}})$ is corresponding to $p_\theta(\bm{x}_0 | \bm{x}_1)$, which has been already included in Eq.~(\ref{eq_ori_vlb}). Therefore, after adding the embedding-rounding step, the VLB of our model is derived as
\begin{align}\label{eq:vlb}
  \mathcal{L}_{\text{VLB}} &= 
  \mathop{\mathbb{E}}_{q_\phi(\bm{x}_{0} | \bm{v})}\left[
    \mathcal{L}_{\text{VLB}}^\text{ori} (\bm{x}_0)
    + \log \frac{q_\phi(\bm{x}_0 | \bm{v})}{p_\theta(\bm{v} | \bm{x}_0)}
  \right] \notag \\
  &= \mathop{\mathbb{E}}_{q_\phi(\bm{x}_{0:T} | \bm{v})}\left[
    \log \frac{q(\bm{x}_T | \bm{x}_0)}{p(\bm{x}_T)}
    + \sum_{t=2}^T \log \frac{q(\bm{x}_{t-1} | \bm{x}_t,\bm{x}_0)}{p_\theta(\bm{x}_{t-1}|\bm{x}_t)} \right. \notag \\
    & + \left. 
    \log \frac{q_\phi(\bm{x}_0 | \bm{v})}{p_\theta(\bm{v} | \bm{x}_0)p_\theta(\bm{x}_0 | \bm{x}_1)}
  \right].
\end{align}
\revised{Following Ho et al.~\cite{ho2020denoising} and like Eq.~(\ref{eq_ori_simple}), $\mathcal{L}_{\text{VLB}}$ can also be simplified as  mean-squared errors. In our case, we re-parameterize $\bm{\mu}_\theta$ as a denoising network $f_\theta(\bm{x}_t,t)$ to predict $\bm{x}_0$, which is modeled as a Transformer decoder as shown in Figure~\ref{fig_pipeline}~(c).} Then, the training objective is simplified to 
\begin{align}\label{eq_simple}
  \mathcal{L}_{\text{simple}} &= \mathop{\mathbb{E}}_{q_\phi(\bm{x}_{0:T} | \bm{v})}\left[
  \sum_{t=2}^T \| \bm{x}_0 - f_\theta(\bm{x}_t,t)\|^2 
  \right. \notag \\
  & + \|\text{Emb}_\phi(\bm{v}) - f_\theta(\bm{x}_1,1)\|^2 -\log p_\theta(\bm{v} | \bm{x}_0)  
  \bigg].
\end{align}
\revised{
During training, $\bm{x}_0$ is sampled by the embedding step, \ie, $\bm{x}_0 \sim q_\phi(\bm{x}_0 | \bm{v})$. Then, $\bm{x}_t$ ($t=1,\ldots,T$) is sampled via the forward diffusion process following Eq.~(\ref{eq:q_sample}).
}

\textbf{Introducing conditional Information}. 
To explicitly control the generated result, conditional information is introduced in diffusion models~\cite{ho2021classifier,rombach2022high,preechakul2022diffusion}, guiding the model to learn conditional distribution $p(\bm{x}|\bm{y})$.
%
In our setting, the conditional information includes the visual appearance features of subject-object pairs and the text features of their categories, as shown in Figure~\ref{fig_pipeline}~(c). This information guides the relation embeddings to become more and more semantically meaningful during the denoising process.


Specifically, we use CLIP~\cite{radford2021learning} encode the features.
The visual features are extracted by CLIP's visual encoder where the image patches are from the detected bounding boxes.
They include the features of subject, object, and subject-object union region. 
The text features of object categories are extracted by CLIP's text encoder with a prompt ``a photo of $<$object$>$''. Then, we get the conditional information by concatenating all the visual and text features, resulting in $\bm{y}= \in \mathbb{R}^{N \times d_y}$ where $N$ is the total number of subject-object pairs in the image.

To integrate conditional information from different modalities, we encode $\bm{y}$ by an encoder $\tau_\theta$ and inject it by using cross-attention~\cite{vaswani2017attention,rombach2022high}. Formally, in each layer of the Transformer decoder, the cross-attention is calculated as $\text{Attention}(\bm{Q},\bm{K},\bm{V}) = \text{Softmax}({\bm{QK}^{\rm T}} / {\sqrt{d}})\bm{V}$ (we only show one attention head for brevity), where
\begin{align}\label{eq:crossattn}
  \bm{Q} = \varphi_t(\bm{x}_t) \bm{W}_Q,  ~~ 
  \bm{K} = \tau_\theta(\bm{y}) \bm{W}_K, ~~ 
  \bm{V} = \tau_\theta(\bm{y}) \bm{W}_V. 
\end{align}
Here, $\varphi_t(\bm{x}_t) \in \mathbb{R}^{L\times d} $ is the intermediate representation of $\bm{x}_t$ in the Transformer, $\tau_\theta(\bm{y}) \in \mathbb{R}^{N \times d}$ is the projected conditional features, and $\bm{W}_Q,\bm{W}_K, \bm{W}_V \in \mathbb{R}^{d \times d}$ are learnable weights. \revised{Based on this mechanism, $f_\theta(\bm{x}_t,t)$ can be rewritten as $f_\theta(\bm{x}_t,t,\tau_\theta(\bm{y}))$. As a result, we can train Diff-VRD by substituting $f_\theta(\bm{x}_t,t,\tau_\theta(\bm{y}))$ in the objective functions in Eq.~(\ref{eq:vlb}) and (\ref{eq_simple}), where  $f_\theta$, $\tau_\theta$ and $\text{Emb}_\phi$ are jointly optimized.}

\subsection{Matching Visual Relations to Subject-Object Pairs}\label{sec_grounding}

After generating the relation sequence, we need to match each relation word (or phrase) to a certain subject-object pair, to get the final VRD result as a $<$subject,predicate,object$>$ triplet, as shown in Figure~\ref{fig_pipeline}~(d). This is achieved by finding a bipartite matching between the sequence of relations and the sequence of subject-object pairs, based on their semantic similarities. 

\textbf{Semantic Similarity}. A straightforward way to compute the semantic similarity is using CLIP~\cite{radford2021learning}. 
Specifically, we initialize the embedding function $\text{Emb}_\phi$ with CLIP-encoded text embeddings. Once $\text{Emb}_\phi$ is learned, we compute the cosine similarities between the relation embeddings and the CLIP-encoded visual embeddings of subject-object union regions. 
Formally, the similarity matrix $\bm{S}$ has a shape of $L \times N$ with each item is calculated as
\begin{align}\label{eq:similarity}
  \bm{S}[i,j] = \cos( \text{Emb}_{\phi}(v_i), \bm{y}_\text{so}^{(j)} ), 
\end{align}
where $\cos(\bm{a},\bm{b}) = \bm{a}^{\rm T}\bm{b} / (\|\bm{a}\|\|\bm{b}\|)$ is the cosine similarity, and $\bm{y}_\text{so}^{(j)} \in \mathbb{R}^d$ is the visual embedding of $j$-th subject-object pair (which is a part of the conditional features $\bm{y}$). Intuitively, since the relation embeddings participate in the cross-attention with the projected conditional features $\tau_\theta(\bm{y})$, we can also calculate the similarity as $\cos(\text{Emb}_{\phi}(v_i),\tau_\theta(\bm{y}^{(j)}))$. 
Our experiments (Sec.~\ref{sec:ablation}) empirically show that they achieve similar performance on both T2I retrieval and VRD recall. 


\textbf{Multi-Round Bipartite Matching}. Given the computed similarity matrix $\bm{S}$, our goal is to assign each relation to a subject-object pair based on $\bm{S}$. Let 
\begin{align}\label{eq:vo}
\bm{v}=[v_1,\ldots,v_L] ~\text{and}~\bm{o} = [o_1,\ldots,o_N,\emptyset]
\end{align}
be the sequence of relations and the sequence of subject-object pairs, respectively. $\bm{o}$ is padded with no pair $\emptyset$ to length $L$. 
Usually, we have $L \geq N$, since one subject-object pair can have multiple relations. We perform bipartite matching for multiple rounds, where each round returns a one-to-one assignment. For example, for the first round, we search for a permutation of $L$ elements $\hat{\sigma}$ by optimizing the cost:
\begin{align}\label{eq_bipartite}
   \hat{\sigma} = \mathop{\arg\min}_{\sigma} \textstyle{\sum}_{j=1}^{L} C(v_{\sigma(j)},h_j), 
\end{align}
where $C(v_{\sigma(j)},h_j) = - \mathbf{1}_{\{h_j \neq \emptyset\}} \bm{S}[\sigma(j),j]$ and $\mathbf{1}_{\{\cdot\}}$ is an indicator function. This optimization problem can be solved efficiently by the Hungarian algorithm~\cite{munkres1957algorithms}, following DETR~\cite{carion2020end}. After the matching in each round, we remove the assigned relations from $\bm{v}$ and repeat the matching operation until each relation is assigned to a subject-object pair.

\subsection{Auxiliary Matching Supervision}\label{sec_grounding_superv}
In fact, till now the model is trained only under the supervision of image-level visual relation labels. 
In other words, the relation sequence is unordered in the generation stage. In the matching stage, the relations are assigned based on semantic similarities without extra training. 

Thus, to achieve more precise semantic similarities, we add an auxiliary matching loss to further utilize the ground-truth annotations. To do this, we first assign ground-truth relations to detected subject-object pairs based on the bounding box IoU following common VRD methods~\cite{li2024nicest,zhang2021spatially,zhang2022efficient}. Then we construct a binary matching matrix $\bm{M} \in \{0,1\}^{L \times N}$ based on the assigned relation labels, where $\bm{M}$ is zero-padded if the number of ground-truth relations is less than $L$. Finally, we add the following binary cross-entropy (BCE) loss in the generation stage: 
\begin{align}\label{eq_match}
 \mathcal{L}_\text{match} = \mathbb{E}_{t\sim \{0,\ldots,T\}} \left[\text{BCE}(\text{Sigmoid}(\bm{S}_t / \kappa ),\bm{M})\right], 
\end{align}
where $\bm{S}_t \in \mathbb{R}^{L \times N}$ with each item at $(i,j)$ calculated as $\cos(\bm{x}_t^{(i)},\tau_\theta(\bm{y}^{(j)}))$,  and $\kappa$ is the temperature parameter. Note that $\mathcal{L}_\text{match}$ handles various levels of noisy input, which considers the semantic ambiguity and enables more robust similarity predictions. The overall objective function after adding the matching supervision is $\mathcal{L} = \mathcal{L}_\text{simple} + \lambda \mathcal{L}_\text{match}$, where $\lambda$ is a  hyperparameter. We show that it helps the model improve the VRD detection recall in ablation studies (\cf~Sec.~\ref{sec:ablation}). 



\section{Experiments}
\subsection{Datasets and Implementation Details}
\textbf{Datasets}. We evaluated our model on both human-object detection (HOI) and scene graph generation (SGG) tasks. For HOI, we used HICO-DET~\cite{chao2018learning} and V-COCO~\cite{gupta2015visual}. HICO-DET contains 37,633 training and 9,546 test images. It was annotated with 117 interaction categories and 80 object categories. V-COCO is a subset of MS-COCO~\cite{lin2014microsoft}, which contains 2,533 training images, 2,867 validation images, and 4,946 test images. It was annotated with 26 interaction categories and the same 80 object categories (as in MS-COCO). For SGG, we used the Visual Genome (VG)~\cite{krishna2017visual} dataset, which contains 108,073 images. Since the number of instances in most predicate categories is quite limited, we followed the widely-used data splits~\cite{xu2017scene} with ground-truth annotations covering 150  object categories and 50  predicate categories. On the VG dataset, we used 70\% images for training and 30\% images for testing. 


 

\textbf{Evaluation Metrics}. 
For SGG, we evaluated the model on the three sub-tasks following prior works~\cite{xu2017scene,li2023label,li2024nicest}, \ie, Predicate Classification (PredCls), Scene Graph Classification (SGCls), and Scene Graph Generation (SGGen). The results are evaluated by two metrics: 1) \textbf{Recall@K (R@K)}: It calculates the proportion of top-K confident predicted $<$subject,~predicate,~object$>$ triplets that are true positives. A predicted triplet is true positive if there is the same triplet category in the ground truth and the bounding boxes of subject and object have sufficient IoUs (\eg, 0.5) with the ground truth. 2) \textbf{mean Recall@K (mR@K)}: It calculates R@K within each predicate category separately and then averages the R@K for all predicates.

For HOI, a common evaluation metric used in existing HOI works is mean average precision (mAP)~\cite{chao2018learning,zhang2022efficient,wu2023end}. However, our generative VRD allows the model to predict interactions beyond the pre-defined categories in HOI datasets, resulting in many ``false positives'' w.r.t. ground-truth annotations (\ie, reasonable but may not be annotated as ground-truth). This makes the mAP meaningless in our setting. Therefore, we used Recall@K instead.  
%
To cover more reasonable predictions, we extended ground-truth predicate classes with synonyms 
when evaluating the detection recall.

\subsection{Implementation Details}\label{sec_implementation}

\textbf{Vocabulary Construction}. To support the generative VRD, we constructed a large predicate vocabulary $\mathcal{V}$ with all possible verbs for visual relations. We first excluded abstract verbs that cannot be visually depicted, such as ``insist'' or ``recall'', by employing a POS tagger~\cite{bird2009nltk} to collect verbs from image captions. Only verbs that have a frequency greater than 0.5 are considered. This process ensures that we focus on verbs that are visually representable and relevant to the content of the image. In addition, we parsed the captions using the Stanford Scene Graph Parser~\cite{schuster2015generating} and collected all the predicate words. This also brings many reasonable prepositions and phrases into the predicate vocabulary. Consequently, we have a common relation vocabulary $\mathcal{V}$ of 4858 words and phrases, which covers all the annotated relation categories in VRD datasets.



\textbf{Model Details}. We model $f_\theta$ as a Transformer decoder where each layer includes a multi-head self-attention, a multi-head cross-attention, and a feed-forward network (FFN). Each block is connected with layer normalization and residual connection. The hidden dimension of attention layers is 512 (the same as the dimension of CLIP embeddings). The number of decoder layers is 6, and the number of attention heads is 8. The FFN and the condition encoder $\tau_\theta$ are two-layer MLPs with an intermediate dimension of 2048 and ReLU activation.

\textbf{Training Details}. The length of the relation sequence is set to $L=32$, which is usually larger than the total number of relation annotations in an image. For the embedding function, each $\text{Emb}_\phi(v_i)$ is initialized by CLIP-encoded text embedding of $v_i$. The total number of diffusion steps $T$ is 2,000 and the total training steps is 40,000.  We set the loss weight and the sigmoid temperature as $\lambda=1.0$ and  $\kappa=0.05$, respectively. The model is trained on 4 2080Ti GPUs with a batch size of 128 and a learning rate of 1e-4.

During training, we pad the relation sequence to $L$ by pseudo labels. They are collected from the predictions of CLIP, \ie, based on the semantic similarities between visual embeddings of subject-object union regions and text embeddings of relation words. The number of ground-truth subject-object pairs is usually less than $L$ (\cf~Eq.~(\ref{eq:vo})). So zero-padding is used for the visual features of subject-object pairs (\ie, the conditional information), and a corresponding attention mask is used in the cross-attention layers. In case some images have too many objects, the visual features are truncated with a maximum number of $L$.

\textbf{Inference Details}. 
\revised{Given an image, we first use DETR~\cite{carion2020end} to detect object bounding boxes and their categories. Then, our Diff-VRD starts from a Gaussian noise to generate relation embeddings conditioned on the visual and text features of subject-object pairs. In the denoising process, we used DDIM~\cite{song2021denoising} scheduler and set the number of denoising steps as 50. The final prediction is formatted as a list of $<$subject, predicate, object$>$ triplets, where each one is scored by the probability from the rounding step $p_\theta(\bm{v} | \bm{x}_0)$. 
}

\begin{table}[t]
    \centering
    \caption{\revised{Recall of HOI detection on HICO-DET~\cite{chao2018learning}. $^\dagger$ stands for implementations that contradict our design spirit, as discussed in Sec.~\ref{sec:conventionalVRD}.
    The vocabulary of training stands for the reference relation set for the embedding-rounding step, and the GT labels used in our training are still within $\mathcal{C}_r$.} 
    }
    \begin{tabular}{lcccc}
    \hline
    Methods                                             & Vocabulary                                    & \multicolumn{3}{c}{HOI Detection (\%)}                                                              \\
                                                        & Training / Testing                            & R@5                             & R@10                            & R@15                            \\ \hline
    EoID~\cite{wu2023end}         & $\mathcal{C}_r ~/~ \mathcal{C}_r$             & 47.08                           & 57.91                           & 62.51                           \\
    CDN~\cite{zhang2021mining}    & $\mathcal{C}_r ~/~ \mathcal{C}_r$             & 48.84                           & 59.03                           & 63.63                           \\
    UPT~\cite{zhang2022efficient} & $\mathcal{C}_r ~/~ \mathcal{C}_r$             & 52.30                           & 64.35                           & \textbf{69.71}                           \\
    GEN-VLKT~\cite{liao2022gen}   & $\mathcal{C}_r ~/~ \mathcal{C}_r$             & \textbf{53.28}                           & \textbf{64.35}                           & 69.14                           \\ \hline
    GEN-VLKT~\cite{liao2022gen}   & $\mathcal{C}_r ~/~ \mathcal{V}$               & 0.71                            & 1.22                            & 1.59                            \\
    THID~\cite{wang2022learning}  & $\mathcal{C}_r ~/~ \mathcal{V}$               & 1.99                            & 3.32                            & 4.36                            \\
    CLIP                          & $\mathcal{V}_{\text{CLIP}} ~/~ \mathcal{V}$   & 1.21                            & 2.72                            & 3.08                            \\
    CLIP                          & $\mathcal{V}_{\text{CLIP}} ~/~ \mathcal{C}_r$ & 17.08                           & 20.25                           & 22.00                           \\ 
    Ours                                                & $\mathcal{V} ~/~ \mathcal{V}$                 & \textbf{17.28}                           & \textbf{21.52}                           & \textbf{23.49}                           \\
    \graytxt{Ours$^\dagger$}                                                & \graytxt{$\mathcal{V} ~/~ \mathcal{C}_r$}               & \graytxt{25.23}                           & \graytxt{31.50}                           & \graytxt{34.21}                           \\
    \graytxt{Ours$^\dagger$}                                      & \graytxt{$\mathcal{C}_r ~/~ \mathcal{V}$}               & \graytxt{4.92} & \graytxt{5.06} & \graytxt{5.11} \\ \hline
    \end{tabular}
    \label{tab:det:hicodet}
\end{table}

\subsection{Evaluation for Conventional VRD}\label{sec:conventionalVRD}

\textbf{Baseline Method}. 
Since existing VRD methods are all trained under the constraint of dataset-specific relation labels, they are not directly comparable with our Diff-VRD which is trained with the common relation vocabulary $\mathcal{V}$. Thus, we developed a baseline method based on CLIP~\cite{radford2021learning}. Specifically, we use CLIP's visual encoder to extract visual embeddings of subject-object union regions and use its text encoder to extract the text embeddings for all relation words. Then, we classify the relations of each subject-object pair based on the cosine similarities between the visual and text embeddings. We keep the top-10 interactions for each subject-object pair.

\textbf{Comparison with SOTA Methods}. We compared our method with state-of-the-art HOI detection methods including EoID~\cite{wu2023end}, CDN~\cite{zhang2021mining}, UPT~\cite{zhang2022efficient}, GEN-VLKT~\cite{liao2022gen}, and THID~\cite{wang2022learning}. The results are shown in Table~\ref{tab:det:hicodet}. We can observe that conventional HOI methods achieve higher recall than our Diff-VRD and the CLIP baseline.
However, one notable point is that our vocabulary constraints are different, \ie, $\mathcal{C}_r$ versus $\mathcal{V}$. For comparisons on the same testing vocabulary constraint $\mathcal{V}$, we modified some zero-shot HOI models (GEN-VLKT~\cite{liao2022gen} and THID~\cite{wang2022learning}) by equipping them with $\mathcal{V}$.
The results show that our Diff-VRD has a significant improvement over them as well as the CLIP baseline, \eg, 17.28\% \emph{vs}. 1.21\% (from CLIP) and 1.99\% (from THID) on R@5.

For training and testing Diff-VRD using $\mathcal{C}_r$, we provide further discussions as follows:
\begin{itemize}
    \item \textbf{Testing with $\mathcal{C}_r$}. 
    Intuitively, a more precise vocabulary constraint brings a higher recall. For example, when replacing $\mathcal{V}$ with $\mathcal{C}_r$ for our model, R@5 improves greatly from 17.28\% to 25.23\%. However, the goal of our Diff-VRD is to identify more meaningful relations in a generative manner and resolve the semantic ambiguity. So we trained the model with a large vocabulary $\mathcal{V}$, based on which the model learned more diverse relations (as well as various synonyms). Testing with a narrower constraint ($\mathcal{C}_r$) is contrary to our original goal, where the model loses a lot of information. For example, it has poor T2I retrieval recall (\eg, 30.59\% on R@10 in Table~\ref{tab:t2i:hicodet}).

    \item \textbf{Training with $\mathcal{C}_r$}. Training on a narrow vocabulary contradicts our diffusion model's design, which inherently results in low performance, \eg, 4.92\% on R@5 in Table~\ref{tab:det:hicodet}. Specifically, we use an embedding step $x_0 \sim q_\phi (x_0 | v)$ to bridge the discrete word embeddings and the continuous embeddings used in the diffusion process. Here each $x_0$ is sampled from the interaction embedding space, and $v$ can be viewed as anchor points provided by the vocabulary. Thus, when the vocabulary is small (\eg, $|\mathcal{C}_r|=117 << |\mathcal{V}|=4858$), the initial anchor points are very sparse and it hurts the learning of the embedding function $q_\phi$ as well as the whole diffusion model. 
    
\end{itemize}

\subsection{Evaluation for Generative VRD}

\textbf{Proxy Tasks and Metrics}. 
To cover the generated relations that exceed the annotations and evaluate the VRD results more comprehensively, we introduce two proxy tasks for indirect evaluation: 1) \textbf{Text-to-Image (T2I) Retrieval}. We tile the detected relation triplets as caption and use a visual-language model (VLM) to perform zero-shot text-to-image retrieval. It assesses the comprehensiveness of the detected triplets by evaluating their ability to accurately describe the contents of the image. We used Recall@K (K=1,5,10) to evaluate this task.  2) \textbf{SPICE PR Curve}. SPICE~\cite{anderson2016spice} originally measures the $F_1$ score of semantic tuples parsed from caption prediction. Unlike caption being a holistic prediction, the VRD predictions are a list of triplets ranked by a confidence score. So we computed the precision-recall curve (PR Curve) by varying the number of top-k triplets returned. As for target captions, we used the official caption annotations for V-COCO, and used a captioning model OFA~\cite{wang2022ofa} to generate captions for HICO-DET. 
For HOI predictions, we introduced synonyms of ``person''. A predicted triplet is considered a true positive if its predicate and object are in ground-truth, and its subject is one of the synonyms of ``person''. 

\begin{table}[!t]
    \centering
    \addtolength{\tabcolsep}{-2.5pt}
    \caption{\revised{Recall of T2I retrieval on HICO-DET~\cite{chao2018learning} dataset. Ours$^\dagger$ stands for the implementation that contradicts our design spirit, as discussed in Sec.~\ref{sec:conventionalVRD}. The caption of each image was built by tiling the top 10 detected relation triplets.}
    }\label{tab:t2i:hicodet}
    \begin{tabular}{lccccc}
    \hline
    Methods                                             & Vocabulary                                  & Retrieval & \multicolumn{3}{c}{T2I Retrieval (\%)} \\
                                                        & Train / Test                          &        Model                       & R@1         & R@5         & R@10       \\ \hline
    EoID~\cite{wu2023end}         & $\mathcal{C}_r ~/~ \mathcal{C}_r$           & CLIP                          & 5.21        & 19.16       & 30.59      \\
    CDN~\cite{zhang2021mining}    & $\mathcal{C}_r ~/~ \mathcal{C}_r$           & CLIP                          & 6.03        & 19.63       & 32.12      \\
    UPT~\cite{zhang2022efficient} & $\mathcal{C}_r ~/~ \mathcal{C}_r$           & CLIP                          & 7.20        & 22.03       & 33.76      \\
    GEN-VLKT~\cite{liao2022gen}   & $\mathcal{C}_r ~/~ \mathcal{C}_r$           & CLIP                          & 6.85        & 23.21       & 34.34      \\
    \graytxt{Ours$^\dagger$}      & \graytxt{$\mathcal{V} ~/~ \mathcal{C}_r$}   & \graytxt{CLIP}                          & \graytxt{4.92}        & \graytxt{19.92}       & \graytxt{30.59}      \\ \hline
    GEN-VLKT~\cite{liao2022gen}   & $\mathcal{C}_r ~/~ \mathcal{V}$             & CLIP                          & 8.96        & 28.13       & 41.32      \\
    THID~\cite{wang2022learning}  & $\mathcal{C}_r ~/~ \mathcal{V}$             & CLIP                          & 7.91        & 23.79       & 35.69      \\
    \revised{CMMP}~\cite{lei2024exploring}  & \revised{$\mathcal{C}_r ~/~ \mathcal{V}$}             & \revised{X-VLM~\cite{zeng2022multi}}                 & \revised{6.66}        & \revised{20.87}       & \revised{30.36}      \\
    CLIP~\cite{radford2021learning}                                                & $\mathcal{V}_{\text{CLIP}} ~/~ \mathcal{V}$ & X-VLM~\cite{zeng2022multi}                         & \textbf{16.00}       & \textbf{41.32}       & \textbf{57.67}      \\
    Ours                                                & $\mathcal{V} ~/~ \mathcal{V}$               & CLIP                          & 11.13       & 32.00       & 45.07      \\
    Ours                                                & $\mathcal{V} ~/~ \mathcal{V}$               & X-VLM~\cite{zeng2022multi}                         & \textbf{15.12}       & \textbf{39.03}       & \textbf{53.63}      \\ \hline
    \end{tabular}
\end{table}

\begin{figure}[!t]
  \centering
  \includegraphics[width=\linewidth]{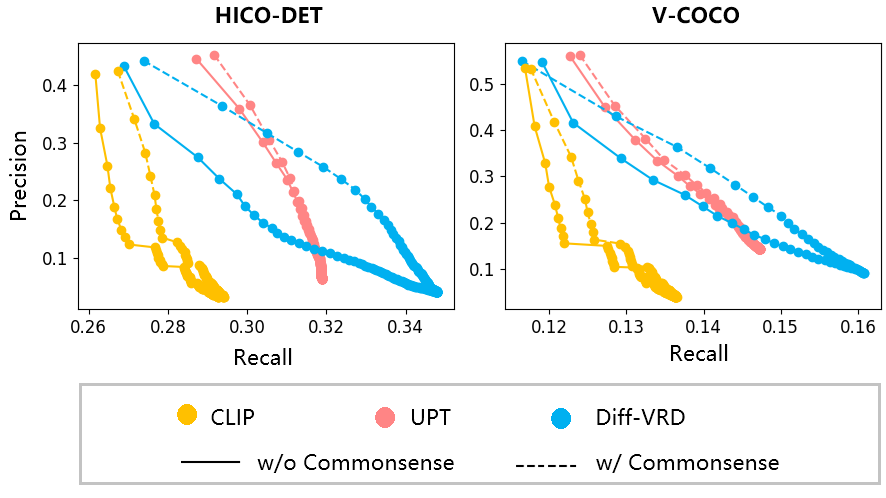}
  \vspace{-3ex}
  \caption{SPICE~\cite{anderson2016spice} precision-recall curve (PR curve)  on HICO-DET~\cite{chao2018learning} and V-COCO~\cite{gupta2015visual} datasets.
  \revised{We used the official caption annotations for COCO and the generated captions from OFA~\cite{wang2022ofa} for HICO-DET.
  The VRD predictions are ranked by their confidence score for SPICE precision and recall evaluation. }
  }\label{fig:spice}
\end{figure}

\begin{figure}[th]
  \centering
  \includegraphics[width=\linewidth]{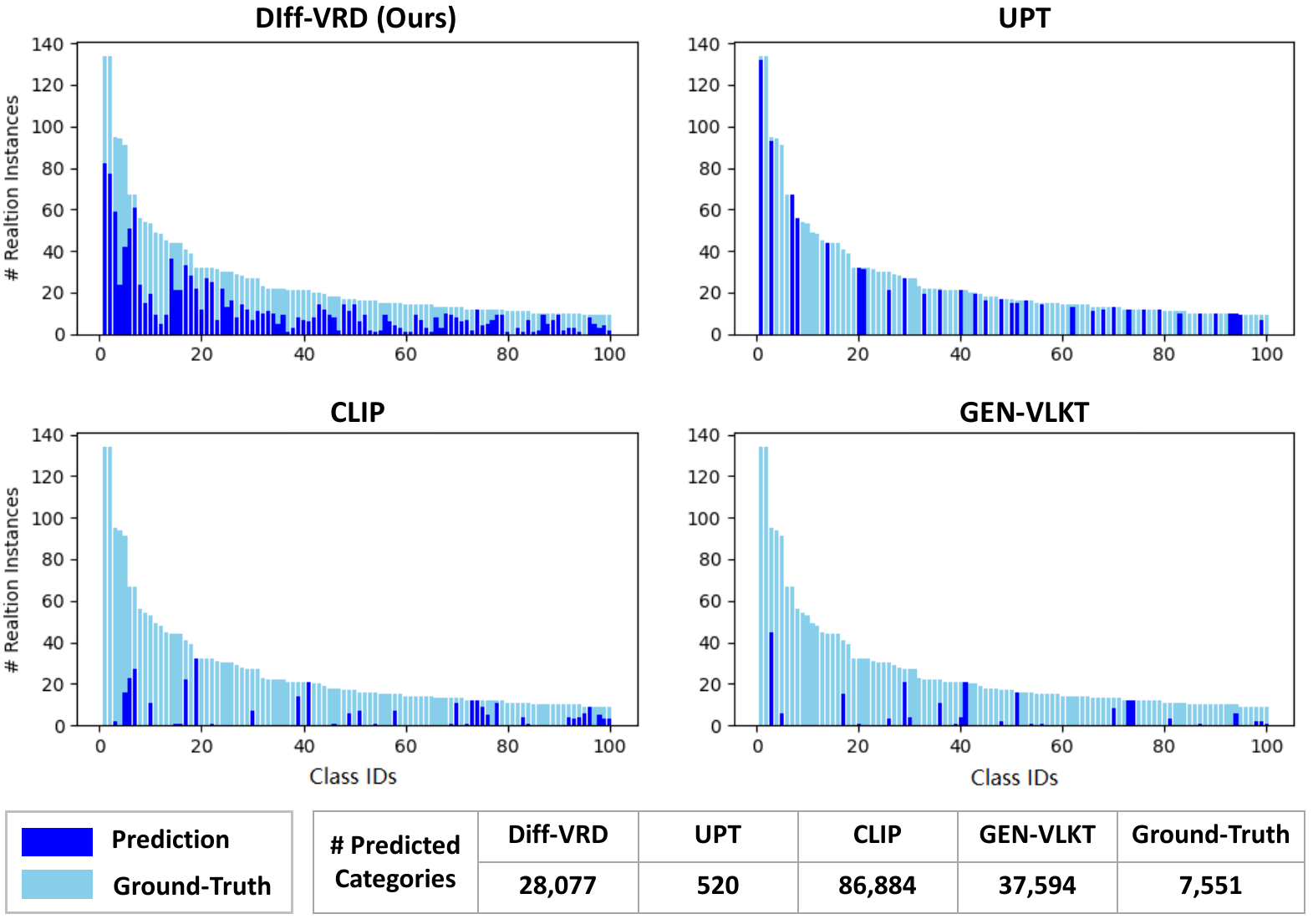}
  \vspace{-3ex}
  \caption{\revised{
  %
  Relation diversity evaluation on HICO-DET~\cite{chao2018learning} test set. Each relation category is a predicate-object combination. The x-axis represents category IDs. The y-axis denotes the total number of relation instances of each category. 
  For a clear visualization, we show only the top 100 categories where at least one of the compared methods hits a ground-truth instance. We also counted the total number of categories predicted by each method, shown at the bottom.
  }}\label{fig:diversity}
\end{figure}

\textbf{T2I Retrieval Evaluation}. We evaluated text-to-image retrieval on HICO-DET and V-COCO datasets (\cf~ Table~\ref{tab:t2i:hicodet} and \ref{tab:t2i:hoi}). 
Firstly, we compared Diff-VRD with SOTA HOI models as shown in Table~\ref{tab:t2i:hicodet}. \revised{We can observe that the T2I recall of Diff-VRD improves when the vocabulary
constraint changes from $\mathcal{C}_r$ to $\mathcal{V}$, \eg, 4.92\%$\rightarrow$11.13\% on R@1. When compared zero-shot HOI models equipped with $\mathcal{V}$ (GEN-VLKT~\cite{liao2022gen}, THID~\cite{wang2022learning} and CMMP~\cite{lei2024exploring}), Diff-VRD still shows clear improvement, \eg, 11.13\% \vs~8.96\% and 7.91\% on R@1.}
In addition, we also compared our model with the CLIP baseline. Since the CLIP's prediction and the T2I retrieval are both conditioned on the similarity between the visual and text embeddings, we utilized another VLM as the retrieval model, \ie, X-VLM~\cite{zeng2022multi}. \revised{We notice that the CLIP baseline shows slightly higher recall than Diff-VRD. This is because CLIP predicts relations solely based on the text-visual similarity ranking, regardless of whether they make sense or not. For example, as shown in Figure~\ref{fig:hicodet}~(a), CLIP predicts ``barbeque'' simply because the image depicts an outdoor cooking activity, where barbeque is a common example of such activities. Therefore, we also conducted the SPICE (\cf~Figure~\ref{fig:spice}) and diversity (\cf~Figure~\ref{fig:diversity}) evaluations for a more comprehensive assessment in the subsequent analysis, where our Diff-VRD shows better overall performance than CLIP.}

\begin{table}[!t]
    \centering
    \caption{Recall (\%) of T2I retrieval on HICO-DET~\cite{chao2018learning} and V-COCO~\cite{gupta2015visual} datasets. $^*$ stands for using the same object detection results.}\label{tab:t2i:hoi}
    \begin{tabular}{lcccccc}
    \hline
    \multicolumn{1}{c}{\multirow{2}{*}{Method}} & \multicolumn{3}{c}{HICO-DET} & \multicolumn{3}{c}{V-COCO} \\
    \multicolumn{1}{c}{}                        & R@1      & R@5     & R@10    & R@1     & R@5     & R@10   \\ \hline
    GT-Object                                   & 4.51     & 16.82   & 27.43   & 7.70    & 24.34   & 34.31  \\
    GT                                          & 7.97     & 24.97   & 37.86   & 8.60    & 26.27   & 37.71  \\
    UPT-Object                                  & 5.55     & 20.57   & 32.53   & 9.51    & 24.68   & 36.91  \\
    UPT$^*$~\cite{zhang2022efficient}                                         & 7.20     & 22.03   & 33.76   & 7.36    & 22.87   & 35.33  \\
    Diff-VRD$^*$                                    & \textbf{11.13}    & \textbf{32.00}   & \textbf{45.07}   & \textbf{10.07}   & \textbf{28.42}   & \textbf{44.05}  \\ \hline
    \end{tabular}
\end{table}

We further investigated the proportion of the predicate's contribution to overall performance, by removing it from the text query and using only object names for T2I retrieval. The results are shown in Table~\ref{tab:t2i:hoi}. Here ``GT'' means using the ground-truth triplet annotations to build the text query, and ``-Object'' means removing all the predicate words. We have the following conclusions: 1) Including predicate words in the text query is necessary and effective for T2I retrieval, \eg, the R@1 of GT improves from 4.51\% to 7.97\% on HICO-DET. 2) When using the same object detection results from a SOTA HOI model, \ie, UPT-Object~\cite{zhang2022efficient}, our Diff-VRD predicts more meaningful and helpful predicate words than UPT, \eg, 11.13\% \vs~7.20\% on R@1 of HICO-DET. 3) Diff-VRD achieved clear improvement over the GT's retrieval results, \eg, 11.13\% \vs~7.97 and 10.07\% \vs~7.70\% on R@1 of HICO-DET and V-COCO, respectively. This means it does predict many reasonable visual relations that are uncovered by the GT annotations.

\textbf{SPICE Evaluation}. We evaluated the precision-recall curve of SPICE~\cite{anderson2016spice} score on the HICO-DET~\cite{chao2018learning} and V-COCO~\cite{gupta2015visual} datasets. We compared Diff-VRD with the CLIP baseline and UPT~\cite{zhang2022efficient}. Results as shown in Figure~\ref{fig:spice}. Here we used a commonsense-based post-processing to filter out unreasonable relation predictions. This technique has been widely adopted in many HOI works~\cite{wu2023end,zhang2021mining,yan2020pcpl,zellers2018neural}, where they use the provided prior from HICO-DET~\cite{chao2018learning} to filter out unreasonable human-object combinations. In our setting, the target triplets (parsed from captions) for SPICE evaluation are not limited to the annotations of datasets. Thus, we used a large language model to construct plausible interactions for each subject-object combination, by masked-language modeling (MLM). Here we use BERT-base~\cite{devlin2018bert} and prompt it by ``a photo of $\bm{s}$ [MASK] $\bm{o}$'', where $\bm{s}$ and $\bm{o}$ are the catogories of subject and object, respectively. Then, it returns a probability distribution $p_{\text{MLM}}(\bm{v})$ over all possible predicate words in the vocabulary\footnote{
the common relation vocabulary $\mathcal{V}$ contains many relations represented as phrases (\eg, ``sit on'', ``in front of''), which are not covered the vocabulary of BERT tokenizer ($\mathcal{V}_{\text{BERT}}$). We map them to tokens in $\mathcal{V}_{\text{BERT}}$ according to the semantic similarities between their embeddings.
}. 
%
%
Consequently, each triplet in the prediction list is re-ranked based on a refined score, \ie, $p_\theta(\bm{v}|\bm{x}) \cdot p_{\text{MLM}}(\bm{v})$.


From the results in Figure~\ref{fig:spice}, we have the following observations: 1) Compared to the CLIP baseline, Diff-VRD exhibits clear and large advantages in terms of both precision and recall. Given the fact that they share the same vocabulary (\ie, $\mathcal{V}$), this justifies the superiority of our Diff-VRD over the CLIP-based classification model. 2) Compared to UPT, Diff-VRD achieves larger improvement after introducing commonsense knowledge. This demonstrates its higher modeling capacity, which is attributed to the diffusion-based design. 3) Diff-VRD achieves the best recall among the three methods, with a slight sacrifice of precision compared to UPT. It also has a larger precision-recall trade-off space.

\revised{
\textbf{Diversity Evaluation.} We further evaluated the diversity of the detected relation categories. We used the relations (predicate-object combinations) parsed in the SPICE task as the ground-truth relations, which cover many more categories than the annotations in the HICO-DET~\cite{chao2018learning} dataset. 
%
%
With results presented in Figure~\ref{fig:diversity}, we have the following observations: 
1) Compared to existing methods---both the closed-set approach UPT~\cite{zhang2022efficient} and the open-vocabulary methods GNE-VLKT~\cite{liao2022gen} and the CLIP~\cite{radford2021learning} baseline, Diff-VRD correctly identifies a larger number of ground-truth relations. 
2) While the open-vocabulary methods also predict a diverse set of categories, their precision is significantly lower, \eg, 37,594 by GNE-VLKT and 86,884 by CLIP (in terms of predicate-object combination categories). They have many incorrect predictions that do not hit the ground truth.
}

\begin{table*}[t]
       \parbox{0.65\linewidth}{
       \centering
       \addtolength{\tabcolsep}{-2pt}
       \caption{\revised{Performance of SGG on VG~\cite{krishna2017visual} test set. $^\dagger$ means the model is implemented on Motif~\cite{zellers2018neural}. Our Diff-VRD is implemented on IEtrans~\cite{zhang2022fine} as an enhancement module. The predicate categories generated by Diff-VRD are based on the vocabulary $\mathcal{V}$ and then mapped to VG's predicate synonyms
       for the closed-set SGG evaluation.}} 
        \label{tab:sgg}
       \revised{\begin{tabular}{lcccccccccccc}
        \hline
        \multirow{3}{*}{Method}               & \multicolumn{4}{c}{PredCls (\%)}          & \multicolumn{4}{c}{SGCls (\%)}                    & \multicolumn{4}{c}{SGGen (\%)}   \\
         & \multicolumn{2}{c}{Recall} & \multicolumn{2}{c}{Mean Recall}   & \multicolumn{2}{c}{Recall}   & \multicolumn{2}{c}{Mean Recall}  & \multicolumn{2}{c}{Recall}          & \multicolumn{2}{c}{Mean Recall}   \\
         & 50 & 100         & 50 & 100 & 50          & 100         & 50 & 100 & 50 & 100 & 50 & 100 \\ \hline
        CLIP~\cite{radford2021learning}                 & 2.1  & 2.9           & 8.4   & 11.3  & 1.3           & 1.7           & 4.4  & 6.4   & 1.2  & 1.5   & 4.4  & 5.3   \\
        VCTree~\cite{tang2019learning}               & 64.5 & 66.5          & 16.3  & 17.7  & 39.3          & 40.2          & 8.9  & 9.5   & 30.2 & 34.6  & 6.7  & 8.0   \\
        Moift~\cite{zellers2018neural}                & 64.0 & 66.0          & 15.2  & 16.2  & 38.0          & 38.9          & 8.7  & 9.3   & 31.0 & 35.1  & 6.7  & 7.7   \\ \hline
        $^\dagger$TDE~\cite{tang2020unbiased}                  & 46.2 & 51.4          & 25.5  & 29.1  & 27.7          & 29.9          & 13.1 & 14.9  & 16.9 & 20.3  & 8.2  & 9.8   \\
        $^\dagger$CogTree~\cite{yu2020cogtree}              & 35.6 & 36.8          & 26.4  & 29.0  & 21.6          & 22.2          & 14.9 & 16.1  & 20.0 & 22.1  & 10.4 & 11.8  \\
        $^\dagger$IEtrans~\cite{zhang2022fine}              & \textbf{48.6} & 50.5          & \textbf{35.8}  & \textbf{39.1}  & 29.4          & 30.2          & \textbf{21.5} & \textbf{22.8}  & \textbf{23.5} & \textbf{27.2}  & \textbf{15.5} & \textbf{18.0}  \\
        +Diff-VRD     & 47.9 & \textbf{52.0} & 33.0  & 38.0  & \textbf{29.5} & \textbf{31.5} & 20.0 & 22.7  & 21.0 & 25.5  & 13.3 & 16.8  \\ \hline
        \end{tabular}}
        }
        \hfill
       \parbox{0.32\linewidth}{
       \centering
       \caption{\revised{Recall of T2I retrieval on VG~\cite{krishna2017visual} test set. The caption of each image was built by tiling the top 10 detected relation triplets (based on their confidence scores). We used X-VLM~\cite{zeng2022multi} for retrieval. 
       }}
       \revised{\begin{tabular}{cccc}
        \hline
        \multirow{2}{*}{Method}           & \multicolumn{3}{c}{T2I Retrieval (\%)} \\
                                          & R@1       & R@5       & R@10      \\ \hline
        CLIP~\cite{radford2021learning}                              & 9.20      & 21.35     & 28.64     \\
        IEtrans~\cite{zhang2022fine}                           & 12.12     & 28.49     & 37.73     \\
        IEtrans+Diff-VRD                          & \textbf{17.72}     & \textbf{36.71}     & \textbf{46.52}     \\ \hline
        \end{tabular}
        \label{tab:t2i:vg}
       }}
\end{table*}

\subsection{Diff-VRD as Enhancement}

Our Diff-VRD can also be applied as an enhancement module for off-the-shelf VRD models. In this case, we first project the detected relations to embeddings via the learned $q_\phi (x_0 | v)$ and corrupt it using $T'$ diffusion steps (we used $T'=250$ in our experiments). Then, we run Diff-VRD on these relation embeddings to generate new relations.
%
We evaluated Diff-VRD on top of a state-of-the-art SGG method IETrans~\cite{zhang2022fine}. The results of SGG and T2I retrieval are shown in Table~\ref{tab:sgg} and Table~\ref{tab:t2i:vg}, respectively. 
\revised{For SGG, since IEtrans is a de-bias SGG method implemented based on a foundation model Moift~\cite{zellers2018neural}, we displayed the results of Moift and another baseline method VCTree~\cite{tang2019learning} in Table~\ref{tab:sgg}. 
We also compared our results to other SOTA models implemented on Moifts, \eg, TDE~\cite{tang2020unbiased}, CogTree~\cite{yu2020cogtree}.

With the results in Table~\ref{tab:sgg}, we can observe that our Diff-VRD, implemented on IEtrans, achieves considerable mean recall improvements over TED and CogTree. On the other hand, the performance compared to IEtrans drops a bit. This is because the ground-truth labels (\ie, the evaluation target) in VG~\cite{krishna2017visual} only have one predicate category for each subject-object pair. However, Diff-VRD generates multiple and more diverse predicates for each pair, which are not necessarily covered by the labels but are still meaningful. To this end, a more comprehensive evaluation is the T2I retrieval task, as shown in Table~\ref{tab:t2i:vg}. Here, Diff-VRD shows a clear improvement over IEtrans, \eg, 17.72\% \vs~12.12\% on R@1. The previous diversity evaluation (\cf~Figure~\ref{fig:diversity}) also demonstrated this point. }
%


\subsection{Ablation Studies}\label{sec:ablation}

\begin{table}[t]
    \centering
    \caption{
    Ablations for matching supervision (M) (\cf~Eq.~(\ref{eq_match})) and different semantic similarities (S) (\cf~Eq.~(\ref{eq:similarity})) on HICO-DET~\cite{chao2018learning} and V-COCO~\cite{gupta2015visual} test sets.
    }
    \begin{tabular}{cccccccc}
        \hline
        \multirow{2}{*}{M} & \multirow{2}{*}{S} & \multicolumn{3}{c}{HOI Detection (\%)}      & \multicolumn{3}{c}{T2I Retrieval (\%)}               \\
                            &                       & R@5              & R@10             & R@15             & R@1              & R@5              & R@10             \\ \hline
        $\times$                   & $\bm{y}_\text{ho}$                 & 17.28          & 21.52          & 23.49          & 11.13          & \textbf{32.00} & \textbf{45.07} \\
        $\times$                   & $\bm{y}$                     & 16.14          & 20.31          & 22.34          & \textbf{11.37} & 31.88          & 44.84          \\
        \checkmark                   & $\bm{y}$                     & \textbf{21.83} & \textbf{25.44} & \textbf{27.12} & 8.67           & 26.49          & 39.03          \\ \hline
    \end{tabular}
    \vspace{-2ex}
    \label{tab:matchsuperv}
\end{table}

\begin{table}[!t]
    \centering
    \caption{\revised{Ablations of HOI detection with different sequence lengths ($L$) on HICO-DET~\cite{chao2018learning} test set. In training, pseudo labels were used for padding when the number of ground-truth relations is less than $L$, \cf~Section~\ref{sec_implementation}.}}
    \begin{tabular}{cccc}
        \hline
        \multirow{2}{*}{$L$} & \multicolumn{3}{c}{HOI Detection (\%)}          \\
                           & R@5              & R@10             & R@15             \\ \hline
        16                 & 21.38          & 24.61          & 25.94          \\
        32                 & 21.83          & 25.44          & 27.12          \\
        48                 & \textbf{22.60} & \textbf{26.62} & \textbf{28.24} \\ \hline
    \end{tabular}
    \vspace{-1ex}
    \label{tab:seqlen}
\end{table}


\textbf{Semantic Similarity and Matching Supervision}. As discussed in Sec.~\ref{sec_grounding} (\cf~Eq.~(\ref{eq:similarity})), the semantic similarity for the matching stage can be computed by interaction embeddings w.r.t. the visual embeddings of subject-object union regions $\bm{y}_\text{so}$, or the projected conditional features $\tau_\theta(\bm{y})$. For the latter, additional matching supervision can be added during training as Eq.~(\ref{eq_match}). We compared these three designs on HICO-DET~\cite{chao2018learning}. From the results in Table~\ref{tab:matchsuperv}, we have the following observations. 1) When removing the matching supervision, $\bm{y}_\text{so}$ achieves slightly better HOI detection recall than $\bm{y}$ while performing similar as $\bm{y}$ in terms of text-to-image retrieval. 2) The matching supervision helps the model achieve better detection recall (\eg, 16.14\% $\rightarrow$ 21.83\% on R@5), but it hurts the text-to-image retrieval recall (\eg, 11.37\% $\rightarrow$ 8.67\% on R@1). This is because the matching information from ground-truth annotations only covers pre-defined interaction categories. It guides the model towards generating more in-domain interactions that satisfy the matching labels while suppressing the diversity. 

\begin{figure}[t]
  \centering
  \includegraphics[width=\linewidth]{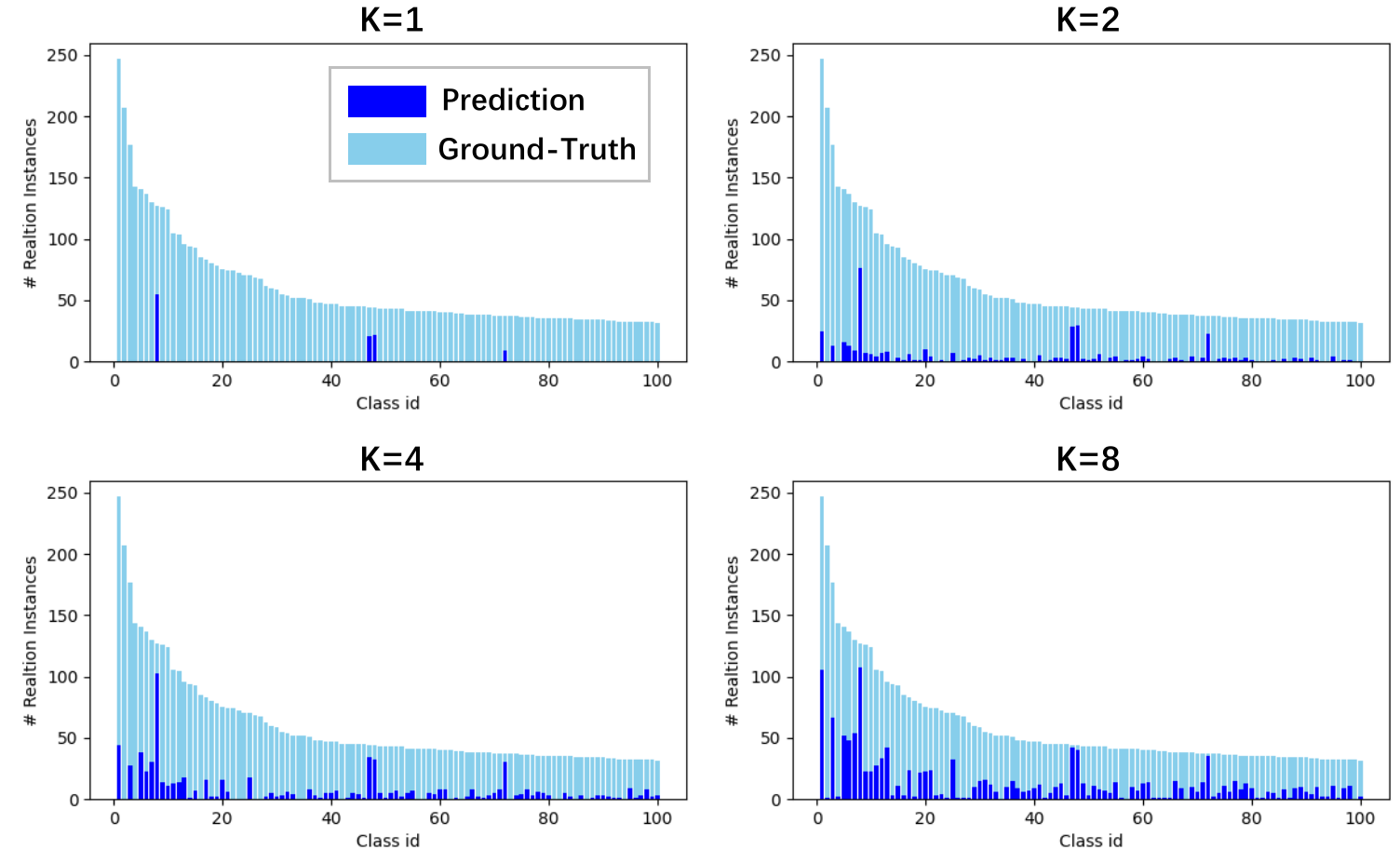}
  \vspace{-3ex}
  \caption{\revised{
  Ablation studies by diversity evaluation for the number of relations ($K$) of each subject-object pair, where Diff-VRD serves as an enhancement on IEtrans~\cite{zhang2022fine}. The experiments were conducted on VG~\cite{krishna2017visual} test set. We used the additionally annotated predicates from~\cite{zhang2022fine} as the ground-truth labels, covering more categories than the original VG dataset.
  }}\label{fig:ablation_k}
\end{figure}
\begin{table}[!h]
\centering
\caption{\revised{Ablations for the number of relations ($K$) of each subject -object pair when using Diff-VRD as an enhancement on IEtrans~\cite{zhang2022fine}. We evaluated PredCls task on VG~\cite{krishna2017visual} test set.}}\label{tab:ablation:k}
\begin{tabular}{ccccc}
\hline
\multirow{2}{*}{$K$} & \multicolumn{2}{c}{Recall} & \multicolumn{2}{c}{Mean Recall} \\
                   & R@50        & R@100        & mR@50           & mR@100          \\ \hline
1                  & \textbf{47.9}        & \textbf{52.0}         & \textbf{33.0}           & \textbf{38.0}           \\
2                  & 47.2        & 52.9         & 31.0           & 37.0           \\
4                  & 46.2        & 53.8         & 28.8           & 35.5           \\
8                  & 44.8        & 54.5         & 26.2           & 33.5           \\ \hline
\end{tabular}
\end{table}

\textbf{Sequence Length}. We investigated the length of interaction sequence (\ie, $L$) on the HICO-DET~\cite{chao2018learning} dataset, and show the results in Table~\ref{tab:seqlen}. We can see that larger $L$ always improves the performance but requires more computational cost. To strike a balance between effectiveness and efficiency, we set $L=32$ for all the experiments.

\revised{
\textbf{Number of Relations for Enhancement}. 
We investigated the number of noisy relation embeddings (denoted as $K$) of each subject-object pair when implementing Diff-VRD as an enhancement. Note that typical SGG models~\cite{zhang2022fine,tang2020unbiased} only predict on predicate category for pair. In our setting, we evaluated different variants with $K=1,2,4,8$ and modeled $\bm{v}$ (\ie, the relation sequence) by all the relations within each subject-object pair. We conducted the diversity evaluation and the PredCls evaluation, as shown in Figure~\ref{fig:ablation_k} and Table~\ref{tab:ablation:k}. In the diversity evaluation, we used additionally annotated predicates from~\cite{zhang2022fine} as the target, covering 1,807 categories, significantly more than the original VG~\cite{krishna2017visual} dataset. 
We can observe that increasing $K$ hurts the conventional SGG metrics (\ie, recall and mean recall). But it can improve the diversity of relation predictions. This is also consistent with our motivation on generalized VRD and the design of Diff-VRD.



}

\subsection{Qualitative Results}

\begin{figure*}[!t]
  \centering
  \includegraphics[width=\linewidth]{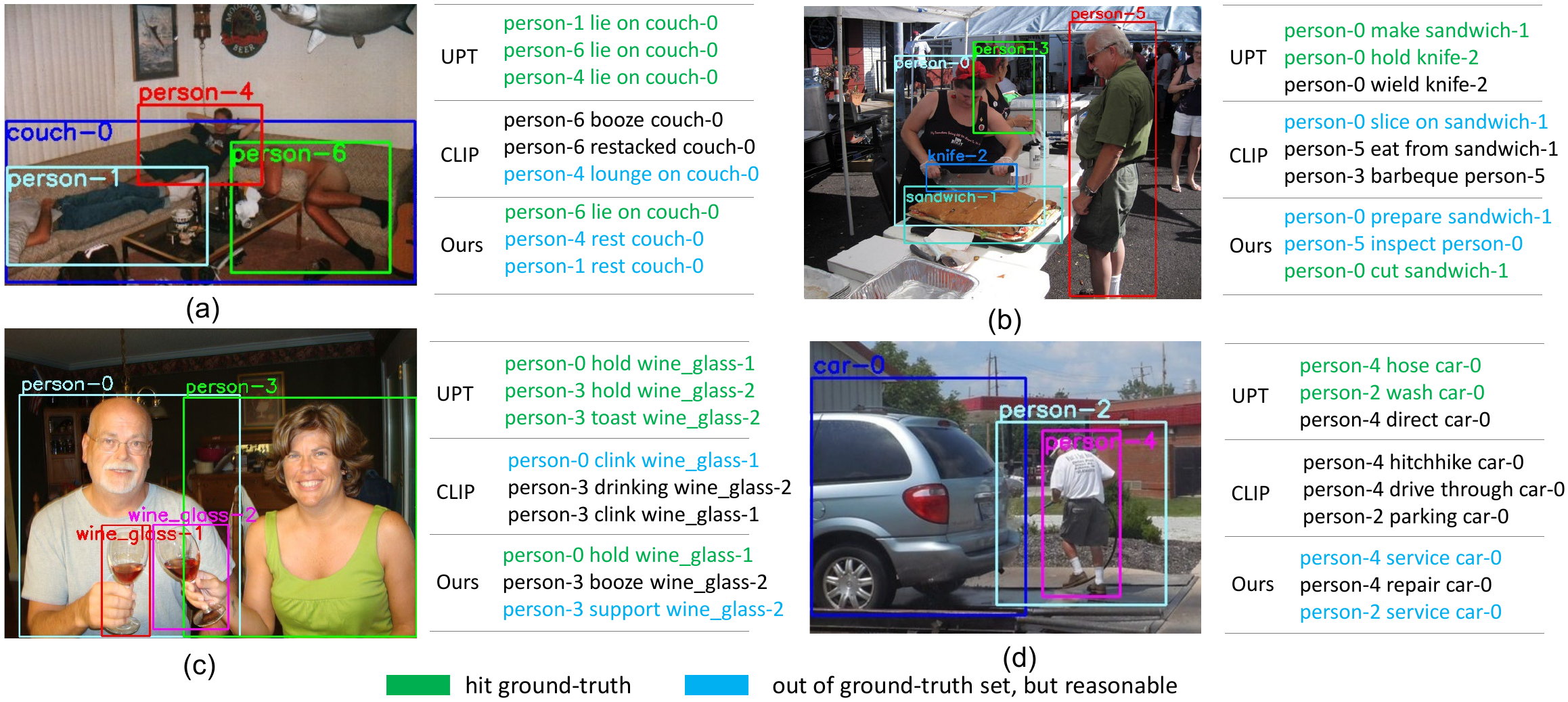}
  \caption{\revised{Qualitative examples of UPT~\cite{zhang2022efficient}, the CLIP~\cite{radford2021learning} baseline, and our Diff-VRD  on HICO-DET~\cite{chao2018learning} dataset. For clarity, we only drew bounding boxes related to the displayed triplets.}}\label{fig:hicodet}
\end{figure*}

\begin{figure}[t]
  \centering
  \includegraphics[width=0.9\linewidth]{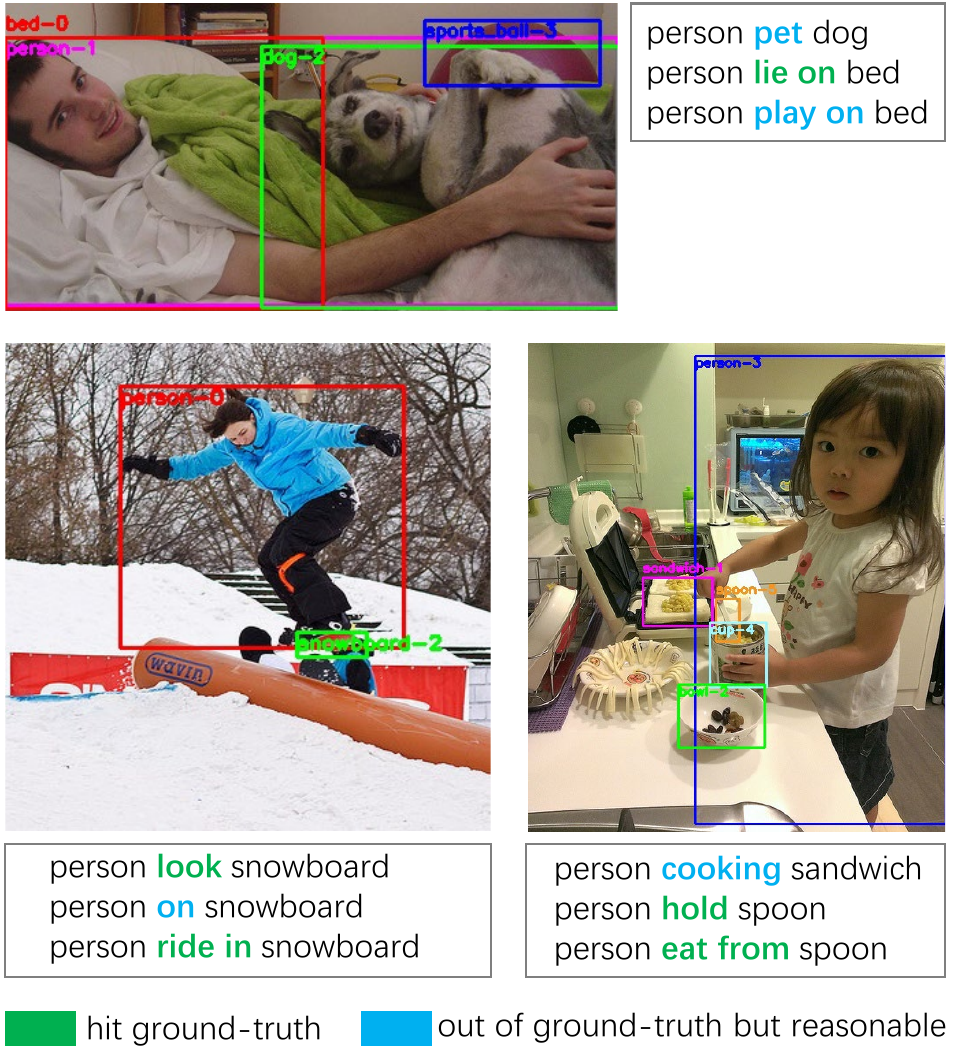}
  \vspace{-2ex}
  \caption{\revised{Qualitative examples of Diff-VRD in V-COCO~\cite{gupta2015visual} dataset. For clarity, we only drew bounding boxes related to the displayed triplets.}}\label{fig:vcoco}
  \vspace{-1ex}
\end{figure}

We show some qualitative examples in Figure~\ref{fig:vcoco} and Figure~\ref{fig:hicodet} for V-COCO~\cite{gupta2015visual} and HICO-DET~\cite{chao2018learning} datasets, respectively. We can observe that our Diff-VRD can detect meaningful and reasonable interactions beyond the pre-defined category, \eg, ``person cooking sandwich'' in Figure~\ref{fig:vcoco} and ``person service car'' in Figure~\ref{fig:hicodet}~(d).

\section{Conclusions and Limitations}

In this paper, we introduce a novel diffusion-based visual relation model, \ie, Diff-VRD. It extends the diffusion-denoising process with an embedding step to project the discrete words to continuous latent embeddings, and a rounding step to round the denoised embeddings to relation words. Diff-VRD achieved generalized VRD by predicting diverse and meaningful relation descriptions that are beyond the ground-truth categories. Two new proxy tasks and metrics are introduced to properly evaluate the generalized VRD predictions, \ie, Text-to-Image (T2I) Retrieval and SPICE PR Curve. We validated the effectiveness of Diff-VRD on both conventional VRD recall and the two proxy tasks. 

\subsection{Limitations}
In some cases, our Diff-VRD makes predictions based on purely visual information rather than interaction information, \eg, ``person booze wine\_glass'' in Figure~\ref{fig:hicodet}~(c). This might be due to the infeasible conditional features which solely focus on objects' visual features while ignoring the context. Our future work will involve modeling improved conditional features and designing enhanced conditional mechanisms for diffusion-based VRD models.

\bibliographystyle{IEEEtran}
\bibliography{ReferenceBib}

\clearpage
\begin{IEEEbiography}[{\includegraphics[width=1in,height=1.25in,clip,keepaspectratio]{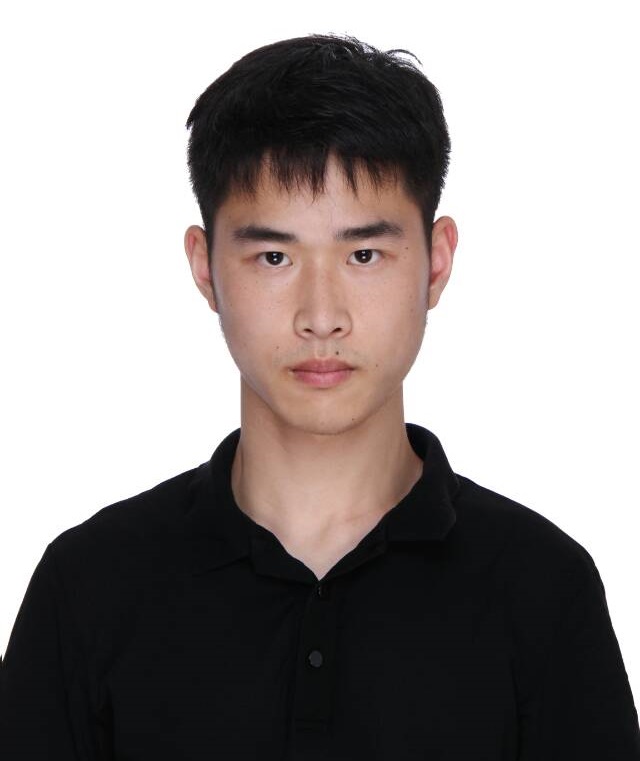}}]{Kaifeng Gao}
received the B.Eng. degree from Zhejiang University, Hangzhou, China, in 2020, where he is currently pursuing the Ph.D. degree with the College of Computer Science. From 2022 to 2023, he was a visiting Ph.D. at Nanyang Technological University and Singapore Management University, Singapore. His current research interests include computer vision, scene analysis and understanding, and video generation.\end{IEEEbiography}
\vspace{-2ex}
\begin{IEEEbiography}[{\includegraphics[width=1in,height=1.25in,clip,keepaspectratio]{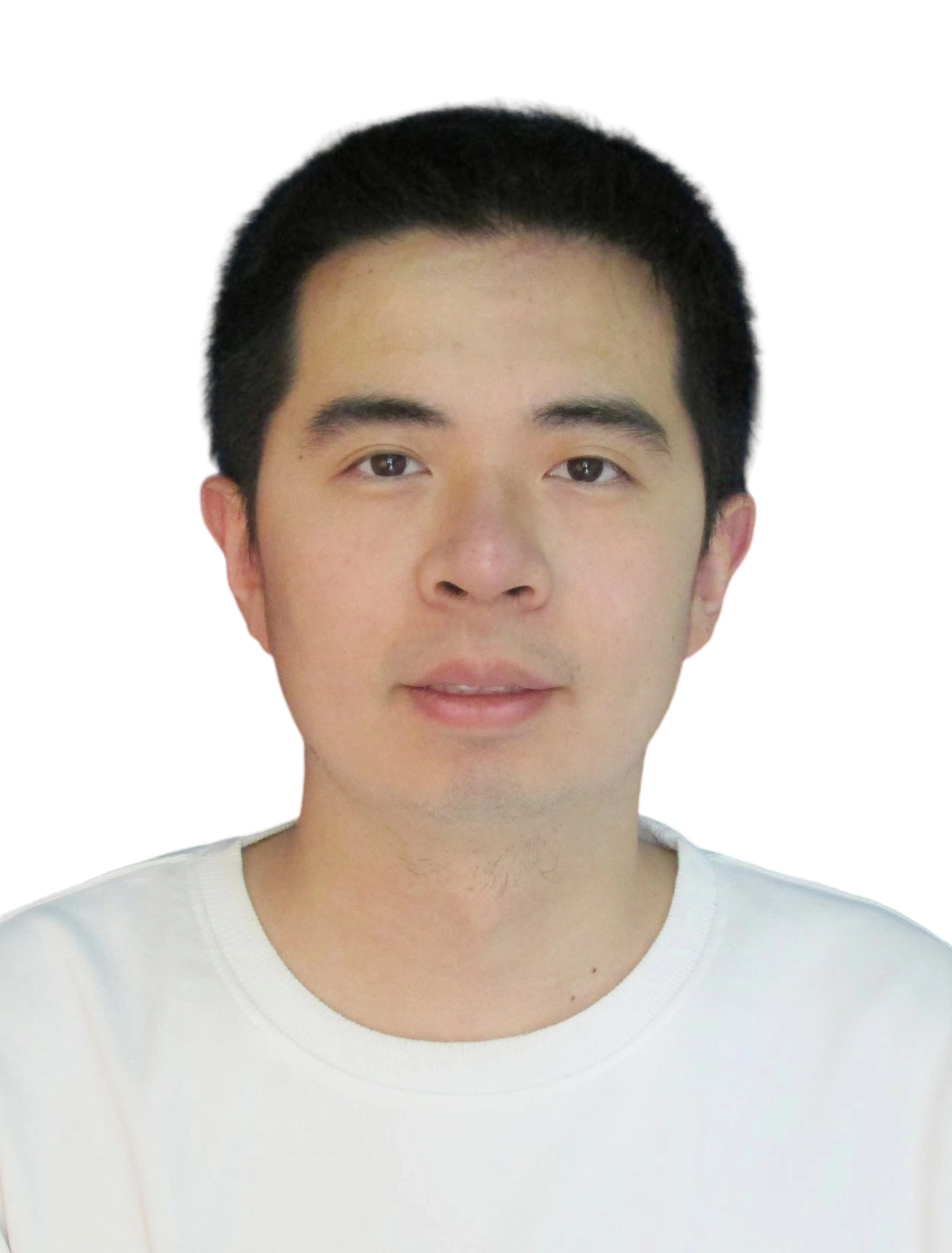}}]{Siqi Chen} received the B.Eng. degree in Software Engineering from Zhejiang University, Hangzhou, China, in 2021, where he subsequently received the master’s degree with the College of Computer Science in 2024. His research interests include computer vision, machine learning, and scene graph generation. \end{IEEEbiography}
\vspace{-2ex}
\begin{IEEEbiography}[{\includegraphics[width=1in,height=1.25in,clip,keepaspectratio]{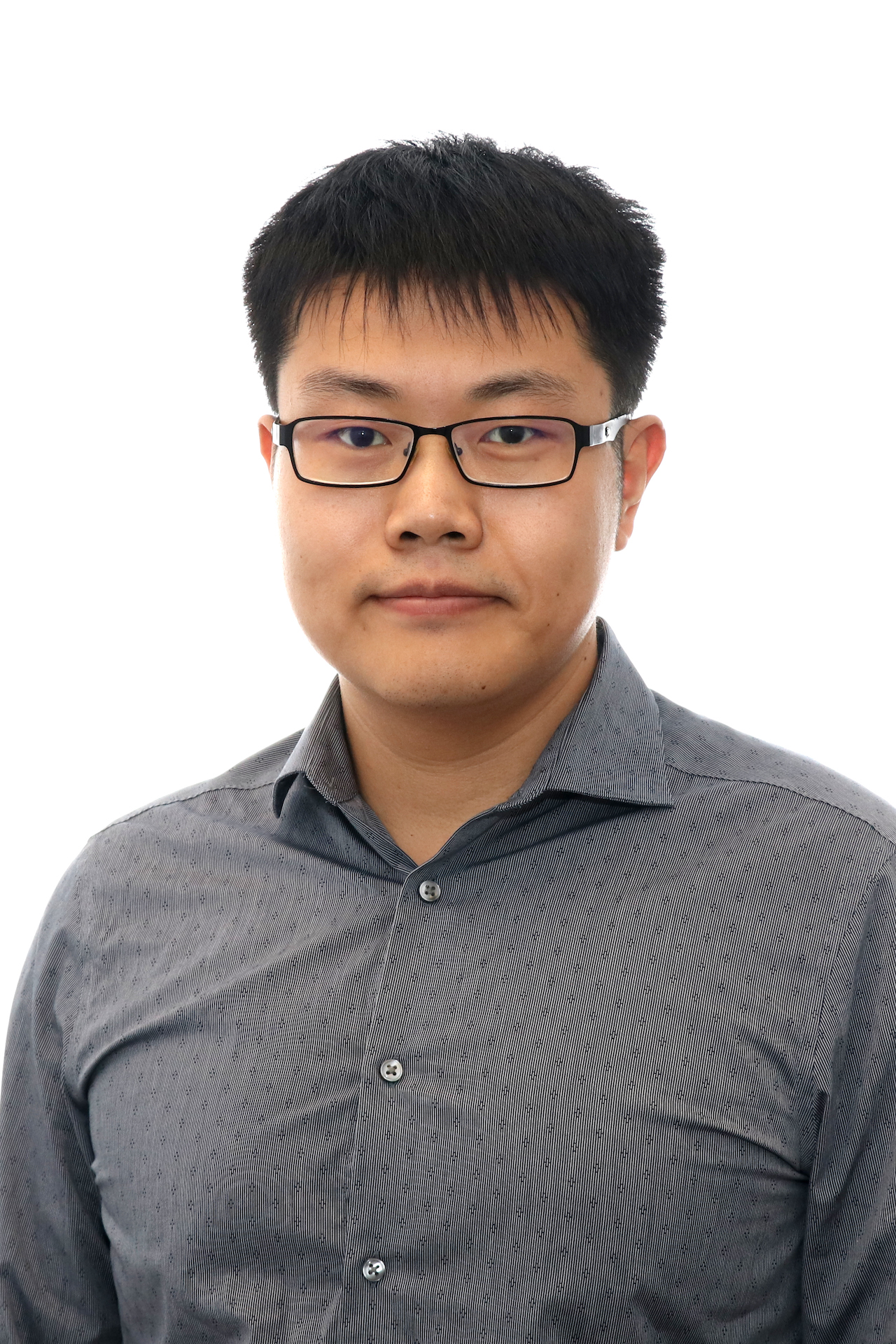}}]{Hanwang Zhang}
received the B.Eng. degree in computer science from Zhejiang University, Hangzhou, China, in 2009, and the PhD degree in computer science from the National University of Singapore, in 2014. He is currently an associate professor with Nanyang Technological University, Singapore. He was a research scientist with the Department of Computer Science, Columbia University, USA. His research interests include computer vision, multimedia, and social media.\end{IEEEbiography}
\vspace{-2ex}
\begin{IEEEbiography}
[{\includegraphics[width=1in,height=1.25in,clip,keepaspectratio]{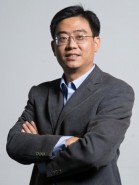}}]{Jun Xiao}
received the Ph.D. degree in computer science and technology from the College of Computer Science, Zhejiang University, Hangzhou, China, in 2007. He is currently a Professor with the College of Computer Science, Zhejiang University. His current research interests include computer animation, multimedia retrieval, and machine learning.\end{IEEEbiography}
\vspace{-2ex}
\begin{IEEEbiography}[{\includegraphics[width=1in,height=1.25in,clip,keepaspectratio]{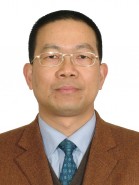}}]{Yueting Zhuang} received his B.Sc., M.Sc. and Ph.D. degrees in Computer Science from Zhejiang University, China, in 1986, 1989 and 1998 respectively. From February 1997 to August 1998, he was a visiting scholar at the University of Illinois at Urbana-Champaign. He served as the Dean of College of Computer Science, Zhejiang University from 2009 to 2017, the director of Institute of Artificial Intelligence from 2006 to 2015. He is now a CAAI Fellow (2018) and serves as a standing committee member of CAAI. He is a Fellow of the China Society of Image and Graphics (2019). Also, he is a member of Zhejiang Provincial Government AI Development Committee (AI Top 30).\end{IEEEbiography}

\begin{IEEEbiography}[{\includegraphics[width=1in,height=1.25in,clip,keepaspectratio]{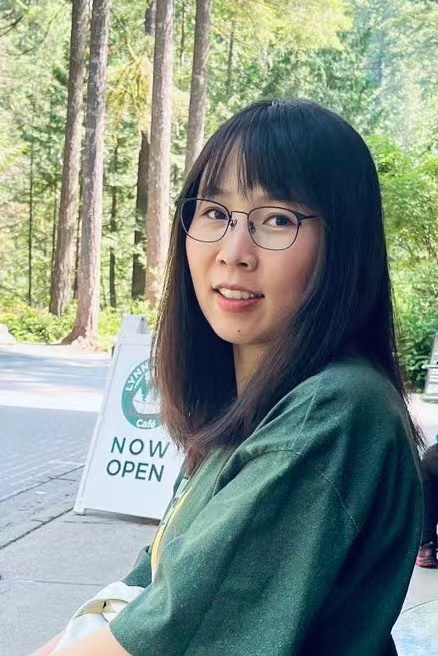}}]{Qianru Sun}
received the Ph.D. degree from Peking University, Beijing, China, in 2016. She is currently an associate professor with the School of Computing and Information System, Singapore Management University, Singapore. Her research interests include computer vision and machine learning that aim to develop efficient algorithms and systems for visual understanding.\end{IEEEbiography}


\clearpage
\onecolumn
\appendix


We first derive the simplified surrogate objective $\mathcal{L}_{\text{simple}}^\text{ori}$ of the original VLB $\mathcal{L}_{\text{VLB}}^\text{ori}$. Then, following the same spirit, we derive $\mathcal{L}_{\text{simple}}$ from $\mathcal{L}_{\text{VLB}}$.

\subsection{$\mathcal{L}_{\text{VLB}}^\text{ori}  \rightarrow \mathcal{L}_{\text{simple}}^\text{ori}$}
We first derive the simplified surrogate objective $\mathcal{L}_{\text{simple}}^\text{ori}$ of the original VLB. 
By decomposing the expectation operation, the $\mathcal{L}_{\text{VLB}}^\text{ori}$ in Eq.~(\ref{eq_ori_vlb}) can be rewritten as:
\begin{align}
  \mathcal{L}_{\text{VLB}}^\text{ori} (\bm{x}_0) &=
  \underbrace{  
  \text{D}_\text{KL}(q(\bm{x}_T | \bm{x}_0) \| p(\bm{x}_T)) 
  }_{\mathcal{L}_T}
  + \sum_{t=2}^T \underbrace{\mathop{\mathbb{E}}_{q(\bm{x}_t | \bm{x}_0)} \left[
      \text{D}_\text{KL}(q(\bm{x}_{t-1} | \bm{x}_t, \bm{x}_0) \| p_\theta(\bm{x}_{t-1} | \bm{x}_t)) 
      \right]
  }_{\mathcal{L}_{t-1}} \notag 
  \\
  &- \underbrace{
    \mathop{\mathbb{E}}_{q(\bm{x}_1 | \bm{x}_0)}\left[
    \log p_\theta(\bm{x}_0 | \bm{x}_1)
    \right]
    }_{\mathcal{L}_0},
\end{align}
where $\text{D}_\text{KL}(q\| p) = \int_{- \infty}^{\infty} q(x) \log \frac{q(x)}{p(x)} \text{d}x$ is the Kullback–Leibler divergence of two distributions.

For $\mathcal{L}_T$, it has no learnable parameters when the variance schedule $\{\beta_t\}$ is fixed.

For $\mathcal{L}_{t-1}$, it can be calculated analytically when both $q(\bm{x}_{t-1} | \bm{x}_t, \bm{x}_0) $ and $p_\theta(\bm{x}_{t-1} | \bm{x}_t)$ are Gaussian. First, following Ho et al.~\cite{ho2020denoising}, we apply a re-parameterization to $\bm{x}_t$. Let $\alpha_t = 1-\beta_t$, $\bar{\alpha}_t = \prod_{i=1}^t \alpha_i$, and $\{\epsilon_t\} \overset{\mathrm{iid}}{\sim} \mathcal{N}(\bm{0},\bm{I})$, we have
\begin{align}\label{eq_re_parameter}
  \bm{x}_t &= \sqrt{\alpha_t}\bm{x}_{t-1} + \sqrt{1-\alpha_t}\epsilon_{t-1} 
  = \sqrt{\alpha_t\alpha_{t-1}}\bm{x}_{t-2} + \sqrt{1-\alpha_t\alpha_{t-1}}\epsilon_{t-2} \notag \\
  &=~\ldots~ = \sqrt{\bar{\alpha}_t}\bm{x}_0 + \sqrt{1-\bar{\alpha}_t}\epsilon_0
\end{align}
Recall that $q(\bm{x}_t | \bm{x}_{t-1}) = \mathcal{N}(\bm{x}_t;\sqrt{1-\beta_t}\bm{x}_{t-1},\beta_t\bm{I})$, and by substituting Eq.~(\ref{eq_re_parameter}) in $q$ with Bayes' rule, we have 
\begin{align}\label{eq_mu_t}
  q(\bm{x}_{t-1} | \bm{x}_t, \bm{x}_0) &= \mathcal{N}(\bm{x}_{t-1}; \tilde{\bm{\mu}}_t(\bm{x}_t,\bm{x}_0),\tilde{\sigma}_t\bm{I}), \notag \\
  \text{where}~\tilde{\bm{\mu}}_t(\bm{x}_t,\bm{x}_0) &= \frac{\sqrt{\bar{\alpha}_{t-1}}\beta_t}{1-\bar{\alpha}_t} \bm{x}_0 + \frac{\sqrt{\alpha_t}(1-\bar{\alpha}_{t-1})}{1-\bar{\alpha}_t}\bm{x}_t,
  ~\text{and}~ \tilde{\sigma}_t = \frac{1-\bar{\alpha}_{t-1}}{1-\bar{\alpha}_{t}}\beta_t.
\end{align}
Then $\mathcal{L}_{t-1}$ can be simplified as 
\begin{align}
  \mathcal{L}_{t-1} = \mathop{\mathbb{E}}_{q(\bm{x}_t | \bm{x}_0)}\left[
    \frac{1}{2 \tilde{\sigma}_t^2} \|\tilde{\bm{\mu}}_t(\bm{x}_t,\bm{x}_0) - \bm{\mu}_\theta(\bm{x}_t,t) \|^2
  \right] + C, ~~ t=2,\ldots,T,
\end{align}
where $C$ is a constant and  $\bm{\mu}_\theta(\bm{x}_t,t)$ is modeled by a neural network to predict $\tilde{\bm{\mu}}_t$.

For $\mathcal{L}_0$, it can also be re-written as a mean-squared error since $p_\theta$ is Gaussian. Specifically,
\begin{align}
  -\mathcal{L}_0 = \mathop{\mathbb{E}}_{q(\bm{x}_1 | \bm{x}_0)}\left[
  \frac{1}{2\sigma_1^2} \|\bm{x}_0 - \bm{\mu}_\theta(\bm{x}_1,1) \|^2 \right]
  + C',
\end{align}
where $C'$ is a constant and $\sigma_1$ is defined  independently. Actually, when $t=1$, the mean of the posterior $q(\bm{x}_{t-1} | \bm{x}_t, \bm{x}_0)$ directly becomes $\bm{x}_0$\footnote{$\tilde{\bm{\mu}}_t$ and $\tilde{\sigma}_t$ are only defined for $t\geq 2$ as in Eq.~(\ref{eq_mu_t}). But we can extend them to $t=1$, where $\tilde{\bm{\mu}}_1$ is acutally $\bm{x}_0$ and $\tilde{\sigma}_1$ is treated as an independently defined value $\sigma_1$.}. By combining $\mathcal{L}_{t-1}$ and $\mathcal{L}_0$ and ignoring the weighting factors (\ie, $\frac{1}{2\tilde{\sigma}_t^2}$ and $\frac{1}{2\sigma_1^2}$), we have the simplified objective function:
\begin{align}\label{eq_ori_simple}
  \mathcal{L}_\text{simple}^\text{ori}(\bm{x}_0) = \sum_{t=1}^T
  \mathop{\mathbb{E}}_{q(\bm{x}_t | \bm{x}_0)}\left[
     \|\tilde{\bm{\mu}}_t(\bm{x}_t,\bm{x}_0) - \bm{\mu}_\theta(\bm{x}_t,t) \|^2
  \right].
\end{align}

Although Eq.~(\ref{eq_ori_simple}) is not a valid lower bound, previous works~\cite{ho2020denoising,song2020improved,dhariwal2021diffusion,li2022diffusion} found it beneficial to sample quality (and also simpler to implement). 

\subsection{$\mathcal{L}_{\text{VLB}} \rightarrow \mathcal{L}_{\text{simple}}$}

In a similar spirit, we can derive $\mathcal{L}_{\text{simple}}$ from $\mathcal{L}_{\text{VLB}}$. First, the $\mathcal{L}_{\text{VLB}}$ can be rewritten as 
\begin{align}
  \mathcal{L}_{\text{VLB}} &= \mathop{\mathbb{E}}_{q_\phi(\bm{x}_{0} | \bm{v})}\left[
    \mathcal{L}_{\text{VLB}}^\text{ori} (\bm{x}_0)
    + \log \frac{q_\phi(\bm{x}_0 | \bm{v})}{p_\theta(\bm{v} | \bm{x}_0)}
  \right] \notag 
  \\ 
  & = \mathop{\mathbb{E}}_{q_\phi(\bm{x}_{0} | \bm{v})}\left[
        \text{D}_\text{KL}(q(\bm{x}_T | \bm{x}_0) \| p(\bm{x}_T)) 
  + \sum_{t=2}^T \mathop{\mathbb{E}}_{q(\bm{x}_t | \bm{x}_0)} \left[
      \text{D}_\text{KL}(q(\bm{x}_{t-1} | \bm{x}_t, \bm{x}_0) \| p_\theta(\bm{x}_{t-1} | \bm{x}_t)) 
      \right]
  \right] \notag
  \\
  &\qquad + \mathop{\mathbb{E}}_{q_\phi(\bm{x}_{0} | \bm{v})}\left[
    \log \frac{q_\phi(\bm{x}_0 | \bm{v})}{p_\theta(\bm{v} | \bm{x}_0)}
    - \mathop{\mathbb{E}}_{q(\bm{x}_1 | \bm{x}_0)}\left[
      \log p_\theta(\bm{x}_0 | \bm{x}_1)
      \right]
    \right] \notag 
  \\ 
  & = \underbrace{
  \mathop{\mathbb{E}}_{q_\phi(\bm{x}_{0} | \bm{v})}\left[
    \text{D}_\text{KL}(q(\bm{x}_T | \bm{x}_0) \| p(\bm{x}_T)) 
    \right]
  }_{\mathcal{L}_T}
  + \sum_{t=2}^T \underbrace{\mathop{\mathbb{E}}_{q_\phi(\bm{x}_t,\bm{x}_0 | \bm{v})} \left[
    \text{D}_\text{KL}(q(\bm{x}_{t-1} | \bm{x}_t, \bm{x}_0) \| p_\theta(\bm{x}_{t-1} | \bm{x}_t)) 
    \right]}_{\mathcal{L}_{t-1}} \notag 
   \\
   &\qquad + \underbrace{\mathop{\mathbb{E}}_{q_\phi(\bm{x}_{1},\bm{x}_{0} | \bm{v})}\left[
    \log \frac{q_\phi(\bm{x}_0 | \bm{v})}{p_\theta(\bm{x}_0 | \bm{x}_1)}
    \right]
   }_{\mathcal{L}_0}
    - \underbrace{\mathop{\mathbb{E}}_{q_\phi(\bm{x}_0 |\bm{v})}\left[
      \log p_\theta(\bm{v} | \bm{x}_0) 
      \right]}_{\mathcal{L}_\text{round}}
\end{align}
Similarily, $\mathcal{L}_T$ has no learnable parameters and can be ignored. $\mathcal{L}_{t-1}$ can be simplified to a mean squared error as $\|\tilde{\bm{\mu}}_t(\bm{x}_t,\bm{x}_0) - \bm{\mu}_\theta(\bm{x}_t,t) \|^2$ with a constant scaling factor. $\mathcal{L}_0$ can also be rewritten as a mean squared error (ignoring the scaling factor) as
\begin{align}
  \mathop{\mathbb{E}}_{q_\phi(\bm{x}_0 |\bm{v})}\left[
    \mathop{\mathbb{E}}_{\bm{x}_0 \sim q_\phi}[\bm{x}_0 | \bm{v}] - \mathop{\mathbb{E}}_{\bm{x}_0 \sim p_\theta}[\bm{x}_0 | \bm{x}_1]
      \right]
    = \mathop{\mathbb{E}}_{q_\phi(\bm{x}_0 |\bm{v})}[\|\text{Emb}_\phi(\bm{v}) - \bm{\mu}_\theta(\bm{x}_1,1) \|^2].
\end{align}
For the last term $\mathcal{L}_\text{round}$, it is a negative log-likelihood with $p_\theta(\bm{v} | \bm{x}_0) = \prod_{i=1}^L p_\theta(v_i | \bm{x}_0^{(i)})$, where $\bm{x}_0^{(i)} \in \mathbb{R}^d$ is the $i$-th item of $\bm{x}_0$, and $p_\theta(v_i | \bm{x}_0^{(i)})$ is modeled as softmax distribution (as introduced in Sec.~\ref{sec_vlb}). Therefore, the overall simplified objective is 
\begin{align}
  \mathcal{L}_{\text{simple-}\mu} = \sum_{t=2}^T \mathop{\mathbb{E}}_{q_\phi(\bm{x}_t,\bm{x}_0 | \bm{v})}\left[
    \|\tilde{\bm{\mu}}_t(\bm{x}_t,\bm{x}_0) - \bm{\mu}_\theta(\bm{x}_t,t) \|^2
    \right]  
  + \mathop{\mathbb{E}}_{q_\phi(\bm{x}_0 |\bm{v})}\left[
    \|\text{Emb}_\phi(\bm{v}) - \bm{\mu}_\theta(\bm{x}_1,1) \|^2
    - \log p_\theta(\bm{v} | \bm{x}_0) 
  \right]. 
\end{align}
In our implementation, we re-parameterize $\mathcal{L}_{\text{simple-}\mu}$ to predict $\bm{x}_0$ by $f_\theta(\bm{x}_t,t)$ in every term (\ie, $\bm{\mu}_\theta(\bm{x}_t,t) = \frac{\sqrt{\bar{\alpha}_{t-1}}\beta_t}{1-\bar{\alpha}_t} f_\theta(\bm{x}_t,t)+ \frac{\sqrt{\alpha_t}(1-\bar{\alpha}_{t-1})}{1-\bar{\alpha}_t}\bm{x}_t$), and then $\|\tilde{\bm{\mu}}_t(\bm{x}_t,\bm{x}_0) - \bm{\mu}_\theta(\bm{x}_t,t) \|^2 \propto \|\bm{x}_0 - f_\theta(\bm{x}_t,t) \|^2$ with a constant scaling factor. Consequently, the final form of $\mathcal{L}_{\text{simple}}$ is~\footnote{Compared to the form in Eq.~(\ref{eq_simple}), here we use a separate (and also more specific) notation of expectations.}
\begin{align}
  \mathcal{L}_{\text{simple}} = \sum_{t=2}^T \mathop{\mathbb{E}}_{q_\phi(\bm{x}_t,\bm{x}_0 | \bm{v})}\left[
    \| \bm{x}_0 - f_\theta(\bm{x}_t,t)\|^2
    \right] 
  + \mathop{\mathbb{E}}_{q_\phi(\bm{x}_0 |\bm{v})}\left[
    \|\text{Emb}_\phi(\bm{v}) -f_\theta(\bm{x}_1,1) \|^2
    - \log p_\theta(\bm{v} | \bm{x}_0) 
  \right]. 
\end{align}

\end{document}